\documentclass[10pt,twocolumn,letterpaper]{article}

\usepackage{cvpr}              

\usepackage{graphicx}
\usepackage{amsmath}
\usepackage{amssymb}

\usepackage{booktabs}
\usepackage{multirow} 
\usepackage[flushleft]{threeparttable}
\usepackage{cuted}  
\usepackage{capt-of}
\usepackage[export]{adjustbox}  
\usepackage[percent]{overpic}  
\usepackage{url}
\usepackage{xcolor}
\usepackage{bm}
\usepackage{pifont}  
\usepackage{mathtools}

\definecolor{cvprblue}{rgb}{0.21,0.49,0.74}
\usepackage[pagebackref,breaklinks,colorlinks,allcolors=cvprblue]{hyperref}

\newcommand{\cmark}{\ding{51}}%
\newcommand{\xmark}{\ding{55}}%

\newcommand{\best}[1]{\textbf{#1}}

\newcommand{\greencheck}{{\color{green}\cmark}}
\newcommand{\redcross}{{\color{red}\xmark}}

\def\paperID{11168} 
\def\confName{CVPR}
\def\confYear{2025}

\title{SelfSplat: Pose-Free and 3D Prior-Free Generalizable 3D Gaussian Splatting}

\newcommand\CoAuthorMark{\footnotemark[\arabic{footnote}]}
\author{Gyeongjin Kang$^{1}$\thanks{Equal contribution}\quad Jisang Yoo$^{1}$\protect\CoAuthorMark\quad Jihyeon Park$^{1}$\quad Seungtae Nam$^{2}$\quad Hyeonsoo Im$^{3}$\\ Sangheon Shin$^{3}$\quad Sangpil Kim$^{4}$\quad Eunbyung Park$^{2}$\thanks{Corresponding author}\\\\Sungkyunkwan University$^{1}$\quad Yonsei University$^{2}$\quad Hanhwa Systems$^{3}$\quad Korea University$^{4}$\\\\
\url{https://gynjn.github.io/selfsplat/}}

\begin{document}
\maketitle

\begin{abstract}
We propose SelfSplat, a novel 3D Gaussian Splatting model designed to perform pose-free and 3D prior-free generalizable 3D reconstruction from unposed multi-view images. These settings are inherently ill-posed due to the lack of ground-truth data, learned geometric information, and the need to achieve accurate 3D reconstruction without finetuning, making it difficult for conventional methods to achieve high-quality results. Our model addresses these challenges by effectively integrating explicit 3D representations with self-supervised depth and pose estimation techniques, resulting in reciprocal improvements in both pose accuracy and 3D reconstruction quality. Furthermore, we incorporate a matching-aware pose estimation network and a depth refinement module to enhance geometry consistency across views, ensuring more accurate and stable 3D reconstructions. To present the performance of our method, we evaluated it on large-scale real-world datasets, including RealEstate10K, ACID, and DL3DV. SelfSplat achieves superior results over previous state-of-the-art methods in both appearance and geometry quality, also demonstrates strong cross-dataset generalization capabilities. Extensive ablation studies and analysis also validate the effectiveness of our proposed methods.

\end{abstract}

\section{Introduction}
\label{sec:intro}
The recent introduction of Neural Radiance Fields (NeRF)~\cite{mildenhall2021nerf} and 3D Gaussian Splatting (3D-GS)~\cite{kerbl20233dgs} had marked a significant advancement in computer vision and graphics, particularly in 3D reconstruction and novel view synthesis. By training on images taken from various viewpoints, these methods can produce geometrically consistent photo-realistic images, providing beneficial for various applications, such as virtual reality~\cite{xu2023vrnerf, jiang2024vrgs}, robotics~\cite{rashid2023lerftogo, fang2024fusionsense}, and semantic understanding~\cite{zhi2021featurenerf, zhou2024featuregs}. Despite their impressive capability in 3D scene representation, training NeRF and 3D-GS requires a large set of accurately posed images as well as iterative per-scene optimization procedures, which limits their applicability for broader use cases.

To bypass the iterative optimization steps, various learning-based generalizable 3D reconstruction models~\cite{wiles2020synsin, yu2021pixelnerf, du2023widebaseline, charatan2024pixelsplat, szymanowicz2024splatter} have been proposed. These models can predict 3D geometry and appearance from a few posed images in a single forward pass. Leveraging large-scale synthetic and real-world 3D datasets, they used pixel-aligned features to extract scene priors from input images and generate novel views through differentiable rendering methods such as volume rendering~\cite{max95} or rasterization~\cite{lassner2021pulsar}. The generated images are then supervised with ground truth images captured from the same camera poses. While this approach enables 3D scene reconstruction without iterative optimization steps, a key limitation remains are as follows: it relies on calibrated images (with accurate camera poses) for both training and inference, thereby constraining its use with less controlled, ``in-the-wild" images or videos.

Recent efforts have integrated camera pose estimation with 3D scene reconstruction, combining multiple tasks within a single framework. By relaxing the constraint of a posed multi-view setup, pose-free generalizable methods~\cite{chen2023dbarf, smith2023flowcam, hong2023unifying, li2024ggrt} aim to learn reliable 3D geometry from uncalibrated images and generate accurate 3D representations in a single forward pass. While these approaches have demonstrated promising results, they still face significant challenges. For example, \cite{smith2023flowcam} relies on error-prone pretrained flow model for pose estimation, often leading to inaccuracies and performance degradation. \cite{chen2023dbarf, li2024ggrt} achieve impressive results but require a per-scene fine-tuning stage, making them computationally expensive for real-world applications. Furthermore, both \cite{smith2023flowcam} and \cite{chen2023dbarf} inherit the limitation of NeRF-based approaches, demanding substantial computational costs due to the volumetric rendering.

In this work, we present SelfSplat, a novel training framework for pose-free, generalizable 3D representations from monocular videos without pretrained 3D prior models or further scene-specific optimizations. We build upon the 3D-GS representation and leverage the pixel-aligned Gaussian estimation pipeline~\cite{charatan2024pixelsplat, szymanowicz2024splatter}, which has demonstrated fast and high-quality reconstruction results. By integrating 3D-GS representations with self-supervised depth and pose estimation techniques, the proposed method jointly predicts depth, camera poses, and 3D Gaussian attributes within a unified neural network architecture. 

3D-GS, as an explicit 3D representation, is highly sensitive to minor errors in 3D positioning. Even slight misplacements of Gaussians can disrupt multi-view consistency, significantly degrading rendering quality~\cite{charatan2024pixelsplat, chen2024mvsplat}. This makes the simultaneous prediction of Gaussian attributes and camera poses especially challenging. The proposed approach, SelfSplat, mitigates this issue by leveraging the strengths of both self-supervised learning and 3D-GS. Exploiting the geometric consistency inherent in self-supervised learning techniques effectively guides the positioning of 3D Gaussians, leading to improved reconstruction accuracy in the absence of camera pose information. Also, harnessing 3D-GS representation and its superior view synthesis capabilities help enhance the accuracy of camera pose estimation, which would otherwise depend solely on 2D image features derived from CNNs~\cite{godard2019digging, sun2023sc} or Transformers~\cite{rockwell20228, chidlovskii2024self}.

While the proposed method is encouraging, simply combining self-supervised learning with explicit 3D geometric supervision has yielded suboptimal results, particularly in predicting accurate camera poses and generating multi-view consistent depth maps. This often results in misaligned 3D Gaussians and inferior 3D structure reconstructions. To address issues from pose estimation errors, we introduce a matching-aware pose network that incorporates additional cross-view knowledge to improve geometric accuracy. By leveraging contextual information from multiple views, this network improves pose accuracy and ensures more reliable estimates across views. Additionally, to support consistent depth estimation, crucial for accurate 3D scene geometry, we develop a depth refinement network. This module uses estimated poses as embedding features which contains spatial information from surrounding views, to achieve accurate and consistent 3D geometry representations.

Once trained in a self-supervised manner, SelfSplat is equipped to perform several downstream tasks, including (1) pose, depth estimation, and (2) 3D reconstruction, including fast novel view synthesis. 
We demonstrate the efficacy of our method on RealEstate10k~\cite{zhou2018re10k}, ACID~\cite{liu2021acid}, and DL3DV~\cite{ling2024dl3dv} datasets providing higher appearance and geometry quality as well as better cross-dataset generalization performance. Extensive ablation studies and analyses also show the effectiveness of our proposed method. The main contributions can be summarized as follows:

\begin{itemize}
    \item We propose SelfSplat, a pose-free and 3D prior-free self-supervised learner from large-scale monocular videos. 
    \item We propose to unify self-supervised learning with 3D-GS representation, harnessing the synergy of both frameworks to achieve robust 3D geometry estimation.
    \item {To address pose estimation errors and inconsistent depth predictions, we introduce the matching-aware pose network and depth refinement module, which enhance geometry consistency across views, ensuring more accurate and stable 3D reconstructions.}
    \item We have conducted comprehensive experiments and ablation studies on diverse datasets, and the proposed SelfSplat significantly outperforms the previous methods. 

\end{itemize}

\section{Related work}
\label{sec:related_work}

\subsection{Pose-free Neural 3D Representations} 

In the absence of camera pose information, recent efforts have aimed to jointly optimize camera poses and 3D scenes. Starting with optimization-based methods, BARF~\cite{lin2021barf} and subsequent research~\cite{bian2023nopenerf, Fu_2024cfgs, keetha2024splatam} addressed this challenge by training poses along with implicit or explicit scene representations. Also, in a generalizable setting with NeRF representations, FlowCam~\cite{smith2023flowcam} utilized pretrained flow estimation model, RAFT~\cite{teed2020raft}, and find the rigid-body motion between 3D point clouds using the Procrustes algorithm~\cite{choy2020deep}. DBARF~\cite{chen2023dbarf} extended the previous optimization-based method~\cite{lin2021barf} and utilized recurrent GRU~\cite{cho2014gru} network for pose and depth estimation. Based on 3D-GS, several methods~\cite{li2024ggrt, hong2024pf3plat, ye2024noposplat} showed pose-free generalizable method by employing explicit 3D representation. However, these methods face practical limitations due to their reliance on pretrained models~\cite{chen2023dbarf, smith2023flowcam, li2024ggrt, ye2024noposplat}, the need for additional fine-tuning stages~\cite{chen2023dbarf, li2024ggrt}, and the computationally intensive volume rendering process~\cite{chen2023dbarf, smith2023flowcam}, all of which hinder their scalability and efficiency in real-world applications. Also, CoPoNeRF~\cite{hong2023unifying} provides poses and radiance fields estimation at the inference stage, it still requires ground-truth pose supervision during training. In contrast, our method can reconstruct 3D scenes and synthesize novel views from unposed images, mitigating the preceding challenges, and offering a more scalable and efficient solution.

\begin{figure*}[!ht]
    \centering
    \includegraphics[width=1.0\textwidth]{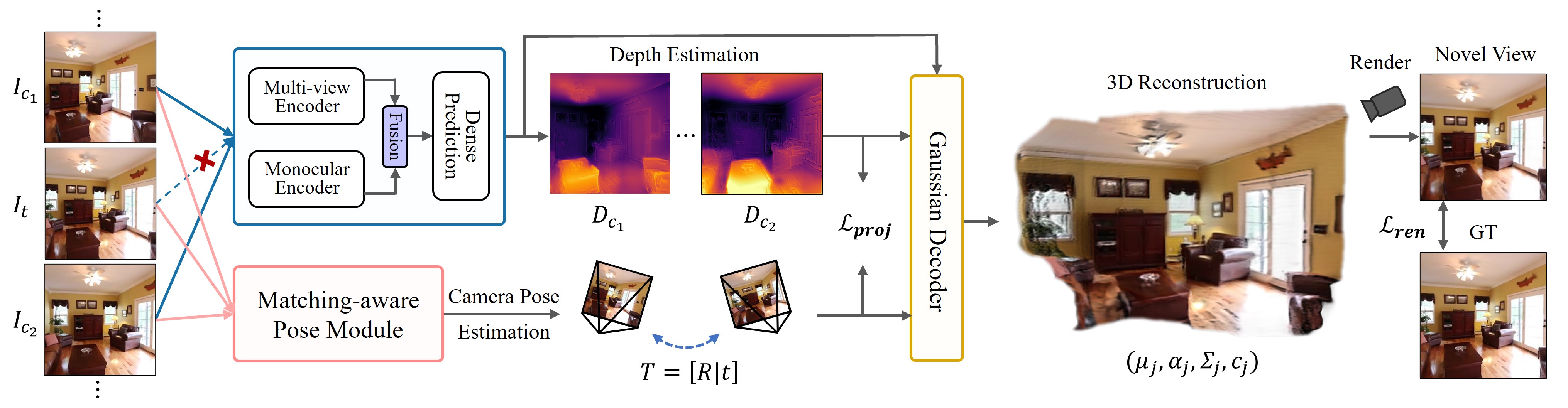}
    
    \vspace{-2mm} \caption{Overview of SelfSplat. Given unposed multi-view images as input, we predict depth and Gaussian attributes from the images, as well as the relative camera poses between them. We unify a self-supervised depth estimation framework with explicit 3D representation achieving accurate scene reconstruction.}
    \vspace{-4mm}
    \label{fig:main}
\end{figure*}

\subsection{Self-supervised Learning for 3D Vision}
Masked Autoencoder (MAE)~\cite{he2022masked,tong2022videomae} is one of self-supervised representation learning framework on video datasets, leveraging their consistency in space and over time. The main objective of MAE is to reconstruct masked patch of pixels or latent features, thereby learning spatiotemporal continuity without any 3D inductive bias. Recently, CroCo~\cite{weinzaepfel2022crocov1, weinzaepfel2023crocov2}, a cross-view completion method which extends previous single-view approaches, has demonstrated a pretraining objective well-suited for geometric downstream tasks, such as optical flow and stereo matching. Expanding on this, DUSt3R~\cite{wang2024dust3r} and MASt3R~\cite{leroy2024grounding} introduce a novel paradigm for dense 3D reconstruction from multi-view image collections.

Another area of self-supervised learning for 3D vision is monocular depth estimation. Without ground-truth depth and camera pose annotations, they utilized the information from consecutive temporal frames using warped image reconstruction as a signal to train their networks. Starting with~\cite{zhou2017unsupervised}, which first introduced the method, and subsequent works~\cite{godard2019digging, bian2019unsupervised, chidlovskii2024naverself} have developed upon this field. In this paper, we also follow the framework of self-supervised depth estimation, but different from previous methods, we combine 3D representation learning, which improves the depth estimation and enables novel view synthesis with resulting 3D scene representations.
\hspace{-1.5mm}
\section{Preliminary}
\label{sec:preliminary}

\subsection{Self-supervised Depth and Pose Estimation}
The self-supervised depth and pose estimation method is a geometric representation learning method from videos or unposed images, which does not require ground-truth depth and pose annotations~\cite{zhou2017unsupervised,yin2018geonet}. Typically, two separate networks are employed for each depth and pose estimation, though these networks may share common representations. Given a triplet of consecutive frames $I_{c_1}, I_{t}, I_{c_2} \in \mathbb{R}^{H \times W \times 3}$, the pose network predicts the relative camera pose between two frames and the depth network produces the depth maps for each frame. While there exist many variants, a typical loss function, $\mathcal{L}_\text{proj}$, to train two networks is 
\hspace{-2mm}
\begin{align}
&\hspace{-1mm}\mathcal{L}_\text{proj} = \texttt{pe}(I_{t}, I_{c_1 \rightarrow t}) + \texttt{pe}(I_{t}, I_{c_2 \rightarrow t}), \\
&\hspace{-1mm}\texttt{pe}(I_a, I_b)\hspace{-1mm} =\hspace{-1mm} \frac{\omega}{2}(1\hspace{-1mm} -\hspace{-1mm} \text{SSIM}(I_a, I_b))\hspace{-1mm} +\hspace{-1mm} (1-\omega)\left\|I_{a}\hspace{-1mm} -\hspace{-1mm} I_{b}\right\|_1,
\label{eq:reproj}
\end{align}
where $I_{c_1 \rightarrow t} \hspace{-1mm} \in \hspace{-1mm} \mathbb{R}^{H \times W \times 3}$ denotes the projected image from $I_{c_1}$ \hspace{-1mm} onto $I_{t}$ using the predicted camera pose and the depth map. $\texttt{pe}(\cdot,\cdot)$ is a photometric reconstruction error, usually calculated using a combination of L1 and SSIM~\cite{ssim} losses, and $\omega$ is a hyperparameter that controls the weighting factor between them~\cite{godard2019digging}.

\subsection{Feed-forward 3D Gaussian Splatting}

Feed-forward 3D Gaussian Splatting methods infer 3D scene structure from input images through a single network evaluation, predicting Gaussian attributes based on pixel-, feature-, or voxel-level tensors. Each Gaussian, defined as $g_j = (\mu_j, \alpha_j, \Sigma_j, c_j)$, includes attributes such as a mean $\mu_j$, a covariance $\Sigma_j$, an opacity $\alpha_j$, and spherical harmonics (sh) coefficients $c_j$.
In particular, our framework adopts a pixel-aligned approach, predicting per-pixel Gaussian primitives along with accurate depth estimations, achieving high-quality 3D reconstruction and fast novel view synthesis. Given multiple input views, the model generates pixel-aligned Gaussians for each image, and combine them to represent the full 3D scene~\cite{charatan2024pixelsplat, szymanowicz2024splatter}.
\vspace{-1mm}
\begin{figure*}[!ht]
    \centering
    \includegraphics[width=1.0\textwidth]{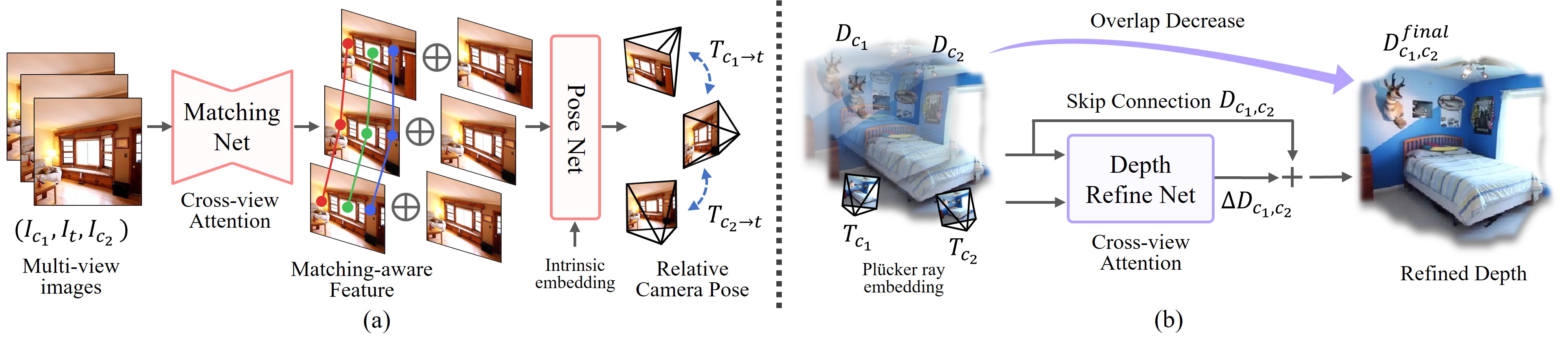}
    \vspace{-6mm}    
    \caption{Matching-aware pose network (a) and depth refinement module (b). We leverage cross-view features from input images to achieve accurate camera pose estimation, and use these estimated poses to further refine the depth maps with spatial awareness.}
    \label{fig:sub}
    \vspace{-4mm}    
\end{figure*}

\section{Methods}
\label{sec:methods}

\subsection{Self-supervised Novel View Synthesis}
\label{sec:overview}

We begin with a triplet of unposed images, $I_{c_1}, I_{t}, I_{c_2} \in \mathbb{R}^{H \times W \times 3}$, which are taken from different viewpoints. Building on the recent pixel-aligned 3D Gaussian Splatting methods, our goal is to predict dense per-pixel Gaussian parameters from input view images,
\begin{equation}
\mathcal{G}_{c_1}, \mathcal{G}_{c_2} = f_{\theta} (I_{c_1}, I_t, I_{c_2}),
\end{equation}
where $f_\theta$ is a feed-forward network with learnable parameters $\theta$, and $\mathcal{G}_{c_1} = \{(\mu_j, \alpha_j, \Sigma_j, c_j)\}_{j=1}^{HW}$ is a generated Gaussians for the input image $I_{c_1}$. Note that we only generate pixel-aligned Gaussians for two input views $I_{c_1}$ and $I_{c_2}$ while excluding the target view $I_t$. This design encourages the network to generalize to novel views $I_t$ during training.
In addition, we train a pose network $f_{\phi}$ to estimate a relative transformation between two images, $T_{c_1 \rightarrow c_2} = f_{\phi}({I}_{c_1}, {I}_{c_2})$, where $T_{c_1 \rightarrow c_2} \in SE(3)$ consists of rotation, $R_{c_1 \rightarrow c_2} \in \mathbb{R}^{3 \times 3}$, and translation, $t_{c_1 \rightarrow c_2} \in \mathbb{R}^{3 \times 1}$, between two images, ${I}_{c_1}$ and ${I}_{c_2}$.
We utilize the estimated camera poses to transform the Gaussian positions in each frame's local coordinate system into an integrated global space. Then, we construct the 3D Gaussian representations for a scene by union of the generated Gaussians as follows,
\begin{equation}
\label{eq:transformation}
\mathcal{G} = \texttt{TR}(\mathcal{G}_{c_1}, T_{c_1 \rightarrow t}) \cup \texttt{TR}(\mathcal{G}_{c_2}, T_{c_2 \rightarrow t}),
\end{equation}
where $\texttt{TR}(\mathcal{G}_{c_1}, T_{c_1 \rightarrow t})$ transforms the generated Gaussian $\mathcal{G}_{c_1}$ into the ${I}_{t}$'s coordinate system, and $\mathcal{G}$ is 
the final 3D Gaussians that are used to render images. The final loss function to jointly train 
both $f_{\theta}$ and $f_{\phi}$ is defined as follows,
\begin{align}
&\hspace{-1mm}\mathcal{L}_\text{total} =\lambda_1\mathcal{L}_\text{proj}+\lambda_2\mathcal{L}_\text{ren}, \\
&\hspace{-1mm}\mathcal{L}_\text{ren}\hspace{-1mm} =\hspace{-3mm} \sum_{I_k \in \{I_{c_1}, I_{c_2}, I_{t}\}} \hspace{-3mm} \gamma_1(1\hspace{-1mm}-\hspace{-1mm}\text{SSIM}(I_k, \hat{I}_k))\hspace{-0.5mm} + \hspace{-0.5mm} \gamma_2 \Vert I_k - \hat{I}_k \Vert_2,
\end{align}
where $\mathcal{L}_{proj}$ is the reprojection loss (\cref{eq:reproj}) and $\mathcal{L}_{ren}$ is the rendering loss that computes the error between input view images, $I_k$, and the rendered images, $\hat{I}_k$, from the constructed Gaussians $\mathcal{G}$. Note that in $\mathcal{L}_{proj}$, we use the rendered depth for ${I}_{t}$ to maintain a consistent scale with estimated depth maps from the context images, $I_{c_1}$ and $I_{c_2}$. In accordance with the prior pose-free generalizable methods, we assume that the camera intrinsic parameters are given from the camera sensor 
metadata~\cite{chen2023dbarf, wang2023pflrm, hong2023unifying, li2024ggrt}.

\subsection{Architecture}
\label{sec:pipeline}

As illustrated in \cref{fig:main}, the proposed SelfSplat consists of four components: a multi-view and monocular encoder, a fusion and dense prediction block, a matching-aware pose estimation network, and a Gaussian decoder.

\noindent \textbf{Multi-view and monocular encoder.}
For multi-view feature extraction from input view images, we begin by processing each image independently through a weight-sharing CNN architecture, followed by a multi-view Transformer to exchange information across different views. Specifically, a ResNet-like architecture~\cite{he2016resnet} is used to extract 4x downsampled features for each view. These features are then refined by a six-block Swin Transformer~\cite{liu2021swin}, which utilizes efficient local window self- and cross-attention mechanisms. The resulting cross-view-aware features are denoted as $F_{c_1}^\text{mv}, F_{c_2}^\text{mv} \in \mathbb{R}^{\frac{H}{4} \times \frac{W}{4} \times C^\text{mv}}$, where $C^\text{mv}$ is the dimension. These features are subsequently processed to generate Gaussian attributes for rendering. As discussed in \cref{sec:overview}, since we do not generate Gaussian attributes for $I_t$, $I_t$ is excluded from the feature extraction in this module.

Despite substantial advancements in multi-view feature matching-based depth estimation methods, such as those leveraging epipolar sampling~\cite{he2020epipolar, charatan2024pixelsplat} or plane-sweep techniques~\cite{yao2018mvsnet, chen2021mvsnerf, chen2024mvsplat}, these approaches continue to face challenges in handling occlusions, texture-less regions, and reflective surfaces. 
To address these limitations, we incorporate a monocular feature extractor, which has demonstrated robust performance across various downstream tasks~\cite{dosovitskiy2020vit, thisanke2023vitsurvey}.
Specifically, we utilize a shared-weight Vision Transformer (ViT) model, CroCo~\cite{weinzaepfel2022crocov1, weinzaepfel2023crocov2}, as a monocular feature extractor.
More specifically, input images are divided into non-overlapping patches with a patch size of 16 and processed by multi-head self-attention blocks and feed-forward networks in parallel.
Then, we obtain robust monocular Transformer features $F_{c_1}^\text{mono}, F_{c_2}^\text{mono} \in \mathbb{R}^{\frac{H}{16} \times \frac{W}{16} \times C^\text{mono}}$, where $C^\text{mono}$ denotes the channel dimension. Similar to the multi-view feature extraction, we do not extract the monocular feature from $I_t$.
It is important to note that, unlike previous methods~\cite{zhang2024transplat, xu2024depthsplat} which use a pretrained DepthAnything~\cite{yang2024depthanything} model as a ViT backbone and thus incorporate 3D priors, we employ CroCov2~\cite{weinzaepfel2023crocov2} weights, allowing us to maintain a fully self-supervised framework

\noindent \textbf{Feature fusion and dense prediction.}
To achieve consistent and fine-grained prediction of Gaussian primitives, we combine the multi- and single-view features, leveraging complementary information from both perspectives to enhance depth accuracy and robustness in complex scenes. We build our feature fusion block with Dense Prediction Transformer (DPT)~\cite{ranftl2021dpt} module. As the spatial resolutions between the two features are different, we first downsample the multi-view features by four to match with monocular ones. Then, CNN-based pyramidal architecture~\cite{lin2017fpn} is adopted to produce features at four different levels. Four intermediate outputs are pulled out from the encoder blocks for the monocular features. These are then simply concatenated at each level and used to produce dense predictions through a combination of reassemble and fusion blocks.

Given the merged features, $F^\text{cat}_{c_1}, F^\text{cat}_{c_2}$, we utilize two branches of dense prediction module, one for the depth of 3D Gaussians, $\texttt{DPT}_\text{depth}$, and the other for the remaining Gaussian attributes $\texttt{DPT}_g$,
\begin{equation}
 \hspace{-1mm}\tilde{D}_k =  \texttt{DPT}_\text{depth}(F^\text{cat}_k), {\tilde{\mathcal{G}}_k} = \texttt{DPT}_g(F^\text{cat}_k),  k \in \{ c_1, c_2\},
\end{equation}
where ${\tilde{\mathcal{G}}_k} = \{(\delta x_j, \delta y_j, \alpha_j, \Sigma_j, c_j)\}_{j=1}^{HW}$ is a set of the predicted Gaussians with all attributes except `$z$' coordinates and $\tilde{D}_k$ is the predicted depth, further processed by the depth refinement module. $\delta x_j$ and $\delta y_j$ are the predicted offsets for each Gaussian, and the Gaussians for each input image in its coordinate system, $\mathcal{G}_{c_1}, \mathcal{G}_{c_2}$, can be obtained by adding the offsets to the pixel coordinates and unprojecting to 3D space using the refined depth $D_{c_1}, D_{c_2}$. Then, we construct the unified Gaussians by transforming them into a target coordinate system with the predicted poses (\cref{eq:transformation}).

\noindent \textbf{Matching-aware pose estimation.}
To enable high-quality rendering and reconstruction, it is crucial to predict accurate camera poses since it defines the transformation in 3D space. We begin by employing the CNN-based pose network from previous studies~\cite{godard2019digging, lai2021video} and introduce our matching-awareness module as a novel encoding strategy. As shown in Fig~\ref{fig:sub}-(a), we use a 2D U-Net~\cite{ronneberger2015unet} with cross-attention blocks to extract multi-view aware features from unposed images.
Unlike the dense prediction module, we also incorporate the target view $I_t$ as input and predict the relative camera poses for $(I_{c_1}, I_{t})$ and $(I_{c_2}, I_{t})$. First, the matching network processes the triplet,
\begin{equation}
F_{c_1}^\text{ma}, F_{t}^\text{ma}, F_{c_2}^\text{ma} =  \texttt{MatchingNet}(I_{c_1}, I_{t}, I_{c_2}),
\end{equation}
where the matching aware features, $F_{k}^\text{ma} \in \mathbb{R}^{H \times W \times 3}$, have the same sizes as input images and inject these features into the pose network. 
\begin{equation}
T_{c_1 \rightarrow t} =  \texttt{PoseNet}( [F_{c_1}^\text{ma}; I_{c_1}; E^\text{int}], [F_{t}^\text{ma}; I_{t}; E^\text{int}]),
\end{equation}
where $[\cdot ; \cdot; \cdot]$ concatenates input tensors along with the channel dimension, and $E^\text{int} \in \mathbb{R}^{H \times W \times 3}$ is ray embedding of camera intrinsic matrix for scale-awareness. More specifically, $E^\text{int}_{x,y} = K^{-1}p(x,y) \in \mathbb{R}^3$, where $K \in \mathbb{R}^{3 \times 3}$ is a camera intrinsic matrix and $p(x,y) \in \mathbb{R}^3$ is a homogeneous coordinate of a pixel coordinate $x,y$. Note that the camera intrinsic parameters vary for different scenes but remain the same across different input views within the same scene.

\noindent \textbf{Pose-aware depth refinement.}
In this module, we refine the estimated depth map, $\tilde{D}_{c_1}, \tilde{D}_{c_2}$, derived from the dense prediction module, $\texttt{DPT}_\text{depth}$, to improve the quality of rendering and reconstruction.
The initial depth estimation, $\tilde{D}_{c_1}, \tilde{D}_{c_2}$, yields inconsistent estimation between input views that negatively impact the overall accuracy of the reconstruction, e.g., incorrectly overlapping Gaussians.
To resolve the limitation, we propose our refine module that leverages cross-view information with spatial awareness.
While a few recent works have proposed depth refinement approaches~\cite{chen2024mvsplat, zhang2024transplat}, our method uniquely differs by utilizing the predicted camera pose as additional information to resolve inconsistencies in the estimated depths across multiple input views. We employs a lightweight 2D U-Net, which takes current depth predictions, input images, and estimated poses as input and outputs residual depths for each view. The operation is defined as follows,
\begin{align}
 \Delta D_{c_1}, \Delta D_{c_2} = \texttt{Refine}( [\tilde{D}_{c_1}; I_{c_1}; E^\text{ext}(T_{c_1 \rightarrow t})], \\
 [\tilde{D}_{c_2}; I_{c_2}; E^\text{ext}(T_{c_2 \rightarrow t})]), \nonumber
\end{align}
where $\Delta D_k$ is the residual for each view depth, and the final depth $D_k = \tilde{D}_k + \Delta D_k$ is obtained by adding the residual to the initial depth estimation for each view. Similar to our pose estimation module, there are cross-attention blocks in U-Net, and we utilize Plücker ray embedding to densely encode our estimated pose into a higher-dimensional representation space, e.g., $E^\text{ext}(T_{c_1 \rightarrow t}) \in \mathbb{R}^{H \times W \times 6}$ (Fig.~\ref{fig:sub}-(b)).

\begin{table}[!ht]
    \centering
    \resizebox{0.7\linewidth}{!}{
    \begin{tabular}{l|ccccc}
     \toprule
     &  \rotatebox[origin=l]{90}{Pose-Free} &  \rotatebox[origin=l]{90}{w/o 3D prior} & \rotatebox[origin=l]{90}{Depth} & \rotatebox[origin=l]{90}{Fast} & \rotatebox[origin=l]{90}{w/o Finetune} \\ \midrule
     VAE~\cite{lai2021video} &     \greencheck         & \greencheck          &    \redcross   & \greencheck & \greencheck      \\
     DBARF~\cite{chen2023dbarf} &     \greencheck     &    \redcross$^2$       & \greencheck      &    \redcross  & \redcross \\
     FlowCAM~\cite{smith2023flowcam}  &         \greencheck     &    \redcross       & \greencheck      &    \redcross  & \greencheck \\
     CopoNeRF~\cite{hong2023unifying} &    \redcross$^1$      &    \greencheck       & \greencheck      &    \redcross  & \greencheck \\ \midrule
     SelfSplat (Ours) & \greencheck  &  \greencheck  &  \greencheck  & \greencheck & \greencheck \\ 
     \bottomrule 
    \end{tabular}}
    \begin{flushleft}
    \end{flushleft}
    \vspace{-5mm}
    \caption{Baseline attributes compared to our proposed method.$^1$ CopoNeRF offers pose-free inference, but requires ground-truth pose supervision during training. $^2$ DBARF is trained from the pretrain generalizable NeRF, IBRNet~\cite{wang2021ibrnet} that has a 3D prior.}
    \vspace{-5mm}
    \label{tab:baselines}
\end{table}

\begin{table*}[!ht]
\renewcommand{\arraystretch}{1}
    \centering
    \resizebox{\linewidth}{!}{
\begin{tabular}{lc|cccccccccccc}
\multicolumn{2}{c|}{}                                                       & \multicolumn{12}{c}{RE10k}                                                                                                                                                                                                        \\ \hline
\multicolumn{1}{c}{}                             &                          & \multicolumn{3}{c}{Average}                            & \multicolumn{3}{c}{Small}                              & \multicolumn{3}{c}{Medium}                             & \multicolumn{3}{c}{Large}                              \\ 
\cmidrule(lr){3-5} \cmidrule(lr){6-8} \cmidrule(lr){9-11} \cmidrule(lr){12-14}
\multicolumn{2}{l|}{Method}                                                 & PSNR $\uparrow$ & SSIM $\uparrow$ & LPIPS $\downarrow$ & PSNR $\uparrow$ & SSIM $\uparrow$ & LPIPS $\downarrow$ & PSNR $\uparrow$ & SSIM $\uparrow$ & LPIPS $\downarrow$ & PSNR $\uparrow$ & SSIM $\uparrow$ & LPIPS $\downarrow$ \\ \hline
\multicolumn{2}{l|}{VAE~\cite{lai2021video}}          & 20.65           & 0.643           & 0.325              & 18.78           & 0.585           & 0.414              & 20.56           & 0.646           & 0.319              & 23.13           & 0.701           & 0.241              \\
\multicolumn{2}{l|}{DBARF~\cite{chen2023dbarf}}       & 12.57           & 0.494           & 0.474              & 10.48           & 0.497           & 0.522              & 12.39           & 0.487           & 0.475              & 15.55           & 0.513           & 0.415              \\
\multicolumn{2}{l|}{FlowCAM~\cite{smith2023flowcam}}  & 22.29           & 0.711           & 0.313              & 20.74           & 0.679           & 0.375              & 22.15           & 0.709           & 0.313              & 24.58           & 0.754           & 0.241              \\
\multicolumn{2}{l|}{CoPoNeRF~\cite{hong2023unifying}} & 21.03           & 0.693           & 0.256              & 19.70          & 0.670           & 0.348              & 20.99           & 0.695           & 0.285              & 22.70           & 0.715           & 0.217              \\
\multicolumn{2}{l|}{Ours}                                                   & \best{24.22}           & \best{0.813}           & \best{0.188}              & \best{21.63}           & \best{0.749}           & \best{0.257}              & \best{24.21}           & \best{0.820}           & \best{0.181}              & \best{27.25}           & \best{0.864}           & \best{0.132}             
\end{tabular}
    }
    \vspace{-2mm}
    \caption{Quantitative results of novel view synthesis on RE10k dataset.}
    \vspace{-2mm}
    \label{tab:quantitative_nvs_re10k} 
\end{table*}

\begin{table*}[!ht]
\renewcommand{\arraystretch}{1}
    \centering
    \resizebox{\linewidth}{!}{
\begin{tabular}{lc|cccccccccccc}
\multicolumn{2}{c|}{}                                                       & \multicolumn{12}{c}{ACID}                                                                                                                                                                                                        \\ \hline
\multicolumn{1}{c}{}                             &                          & \multicolumn{3}{c}{Average}                               & \multicolumn{3}{c}{Small}                              & \multicolumn{3}{c}{Medium}                             & \multicolumn{3}{c}{Large}                              \\
\cmidrule(lr){3-5} \cmidrule(lr){6-8} \cmidrule(lr){9-11} \cmidrule(lr){12-14}
\multicolumn{2}{l|}{Method}                                                 & PSNR $\uparrow$ & SSIM $\uparrow$ & LPIPS $\downarrow$ & PSNR $\uparrow$ & SSIM $\uparrow$ & LPIPS $\downarrow$ & PSNR $\uparrow$ & SSIM $\uparrow$ & LPIPS $\downarrow$ & PSNR $\uparrow$ & SSIM $\uparrow$ & LPIPS $\downarrow$ \\ \hline
\multicolumn{2}{l|}{VAE~\cite{lai2021video}}          & 24.02           & 0.666           & 0.287              & 23.76           & 0.679           & 0.306              & 24.21           & 0.664           & 0.295              & 23.99           & 0.658           & 0.263              \\
\multicolumn{2}{l|}{DBARF~\cite{chen2023dbarf}}       & 15.48           & 0.572           & 0.397              & 13.87           & 0.618           & 0.422              & 15.51           & 0.584           & 0.398              & 16.72           & 0.521           & 0.378              \\
\multicolumn{2}{l|}{FlowCAM~\cite{smith2023flowcam}}  & 25.59           & 0.721           & 0.294              & 25.37           & 0.730           & 0.312              & 25.64           & 0.714           & 0.305              & 25.73           & 0.723           & 0.267              \\
\multicolumn{2}{l|}{{CoPoNeRF~\cite{hong2023unifying}}} & 23.30               & 0.668               & 0.278                  & 22.92               & 0.692               & 0.304                  & 23.42           & 0.667           & 0.283              & 23.45               & 0.649               & 0.251                  \\
\multicolumn{2}{l|}{Ours}                                                   & \best{26.71}           & \best{0.801}           & \best{0.196}              & \best{25.65}           & \best{0.776}           & \best{0.237}              & \best{26.87}           & \best{0.797}           & \best{0.198}              & \best{27.33}           & \best{0.826}           & \best{0.159}             
\end{tabular}
    }
    \vspace{-2mm}
    \caption{Quantitative results of novel view synthesis on ACID dataset.}
    \vspace{-1mm}
    \label{tab:quantitative_nvs_acid}
\end{table*}

\begin{figure*}[!ht]
    \centering
    \includegraphics[width=1.0\textwidth]{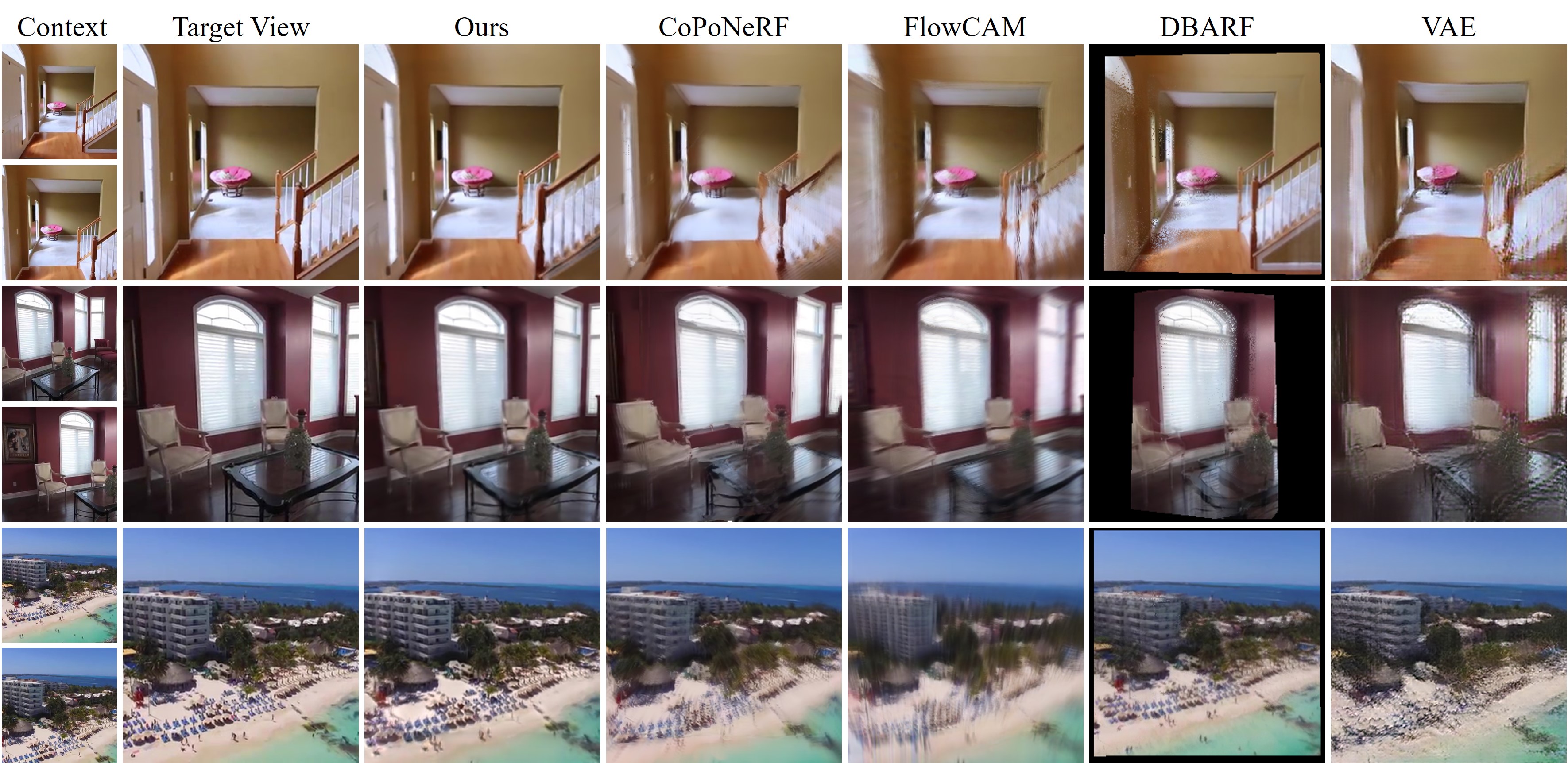}
    \vspace{-5mm}      
    \caption{Qualitative comparison of novel view synthesis on RE10k (top two rows) and ACID (bottom row) datasets.}
    \vspace{-4mm}  
    \label{fig:qual_re10k_acid}
\end{figure*}

\begin{table*}
\renewcommand{\arraystretch}{1}
    \centering
    \resizebox{\linewidth}{!}{
\begin{tabular}{ll|cccccccccccccccc}
\multicolumn{2}{c|}{}                                                       & \multicolumn{16}{c}{RE10k}                                                                                                                                                                                                                                                                                                    \\ \hline
\multicolumn{1}{c}{}                  & \multicolumn{1}{c|}{}               & \multicolumn{4}{c}{Average}                                                   & \multicolumn{4}{c}{Small}                                                     & \multicolumn{4}{c}{Medium}                                                    & \multicolumn{4}{c}{Large}                                                     \\
\cmidrule(lr){3-6} \cmidrule(lr){7-10} \cmidrule(lr){11-14} \cmidrule(lr){15-18}
                                      &                                     & \multicolumn{2}{c}{Rotation}          & \multicolumn{2}{c}{Translation}       & \multicolumn{2}{c}{Rotation}          & \multicolumn{2}{c}{Translation}       & \multicolumn{2}{c}{Rotation}          & \multicolumn{2}{c}{Translation}       & \multicolumn{2}{c}{Rotation}          & \multicolumn{2}{c}{Translation}       \\
\multicolumn{2}{l|}{Method}                                                 & Avg.$^\circ\downarrow$ & Med.$^\circ\downarrow$ & Avg.$^\circ\downarrow$ & Med.$^\circ\downarrow$ & Avg.$^\circ\downarrow$ & Med.$^\circ\downarrow$ & Avg.$^\circ\downarrow$ & Med.$^\circ\downarrow$ & Avg.$^\circ\downarrow$ & Med.$^\circ\downarrow$ & Avg.$^\circ\downarrow$ & Med.$^\circ\downarrow$ & Avg.$^\circ\downarrow$ & Med.$^\circ\downarrow$ & Avg.$^\circ\downarrow$ & Med.$^\circ\downarrow$ \\ \hline
\multicolumn{2}{l|}{VAE~\cite{lai2021video}}          & 3.859             & 2.818             & 115.746             & 113.598             & 6.971             & 6.936             & 118.211             & 116.226             & 3.595             & 3.037             & 112.444             & 109.959             & 1.127             & 0.525             & 123.840             & 132.181             \\
\multicolumn{2}{l|}{DBARF~\cite{chen2023dbarf}}       & 2.471             & 1.471             & 36.069             & 24.308             & 4.771             & 3.009             & 38.292             & 25.822             & 2.043             & 1.440             & 32.749             & 20.899             & 1.222             & 0.914             & 44.500             & 35.788             \\
\multicolumn{2}{l|}{FlowCAM~\cite{smith2023flowcam}}  & 1.438             & 1.160             & 22.524             & 15.917             & 2.483             & 1.741             & 26.044             & 19.583             & 1.264             & 1.137             & 19.889             & 14.355             & 0.801             & 0.745             & 27.179             & 19.328             \\
\multicolumn{2}{l|}{CoPoNeRF~\cite{hong2023unifying}} & 0.839                 & 0.587                 & 9.175                 & 5.538                 & \best{1.281}                 & 0.830                 & \best{10.173}                 & \best{6.047}                 & 0.784             & 0.583             & 8.431             & 5.179             & 0.511                 & 0.326                 & 10.487                 & 6.232                 \\
\multicolumn{2}{l|}{Ours}                                                   & \best{0.750}             & \best{0.362}             & \best{9.095}             & \best{4.438}             & 1.523             & \best{0.701}             & 14.954             & 7.267             & \best{0.603}             & \best{0.346}             & \best{7.593}             & \best{3.976}             & \best{0.344}             & \best{0.206}             & \best{7.281}             & \best{3.799}            
\end{tabular}
    }
    \vspace{-2mm}
    \caption{Quantitative results of pose estimation on RE10k dataset.}
    \vspace{-2mm}
    \label{tab:quantitative_pose_re10k}
\end{table*}

\begin{table*}[!ht]
\renewcommand{\arraystretch}{1}
    \centering
    \resizebox{\linewidth}{!}{
\begin{tabular}{ll|cccccccccccccccc}
\multicolumn{2}{c|}{}                                                       & \multicolumn{16}{c}{ACID}                                                                                                                                                                                                                                                                                                    \\ \hline
\multicolumn{1}{c}{}                  & \multicolumn{1}{c|}{}               & \multicolumn{4}{c}{Average}                                                   & \multicolumn{4}{c}{Small}                                                     & \multicolumn{4}{c}{Medium}                                                    & \multicolumn{4}{c}{Large}                                                     \\
\cmidrule(lr){3-6} \cmidrule(lr){7-10} \cmidrule(lr){11-14} \cmidrule(lr){15-18}
                                      &                                     & \multicolumn{2}{c}{Rotation}          & \multicolumn{2}{c}{Translation}       & \multicolumn{2}{c}{Rotation}          & \multicolumn{2}{c}{Translation}       & \multicolumn{2}{c}{Rotation}          & \multicolumn{2}{c}{Translation}       & \multicolumn{2}{c}{Rotation}          & \multicolumn{2}{c}{Translation}       \\
\multicolumn{2}{l|}{Method}                                                 & Avg.$^\circ\downarrow$ & Med.$^\circ\downarrow$ & Avg.$^\circ\downarrow$ & Med.$^\circ\downarrow$ & Avg.$^\circ\downarrow$ & Med.$^\circ\downarrow$ & Avg.$^\circ\downarrow$ & Med.$^\circ\downarrow$ & Avg.$^\circ\downarrow$ & Med.$^\circ\downarrow$ & Avg.$^\circ\downarrow$ & Med.$^\circ\downarrow$ & Avg.$^\circ\downarrow$ & Med.$^\circ\downarrow$ & Avg.$^\circ\downarrow$ & Med.$^\circ\downarrow$ \\ \hline
\multicolumn{2}{l|}{VAE~\cite{lai2021video}}          &1.631             & 0.461             & 86.180             & 79.987             & 2.789             & 0.617             & 81.056             & 72.060             & 1.568             & 0.497             & 85.372             & 77.761             & 0.806             & 0.355             & 91.187             & 87.526             \\
\multicolumn{2}{l|}{DBARF~\cite{chen2023dbarf}}       & 1.975             & 0.860             & 49.906             & 95.399             & 2.899             & 1.060             & 52.275             & 36.895             & 1.879             & 0.751             & 49.396             & 36.702             & 1.373             & 0.745             & 48.693             & 33.672             \\
\multicolumn{2}{l|}{FlowCAM~\cite{smith2023flowcam}}  & 3.846             & 2.819             & 92.786             & 82.976             & 4.787             & 3.574             & 84.599             & 76.659             & 4.297             & 3.001             & 91.262             & 81.019             & 2.548             & 2.168             & 101.074             & 91.248             \\
\multicolumn{2}{l|}{{CoPoNeRF~\cite{hong2023unifying}}} & 1.058                 & 0.354                 & 19.888                 & 9.757                 & \best{1.731}                 & 0.503                 & \best{20.572}                 & 9.449                 & 1.025             & 0.365             & 19.617             & 9.401             & 0.575                 & 0.261                 & 19.694                 & 10.578                 \\
\multicolumn{2}{l|}{Ours}                                                   & \best{0.981}             & \best{0.199}             & \best{16.329}             & \best{4.535}             & 1.787             & \best{0.362}             & 21.631             & \best{5.681}             & \best{0.952}             & \best{0.219}             & \best{16.006}             & \best{4.518}             & \best{0.386}             & \best{0.134}             & \best{12.592}             & \best{3.828}            
\end{tabular}

    }
\vspace{-2mm}    
\caption{Quantitative results of pose estimation on the ACID dataset.}
\vspace{-1mm}
    \label{tab:quantitative_pose_acid}
\end{table*}

\begin{figure*}[!ht]
    \centering
    \includegraphics[width=1.0\linewidth]{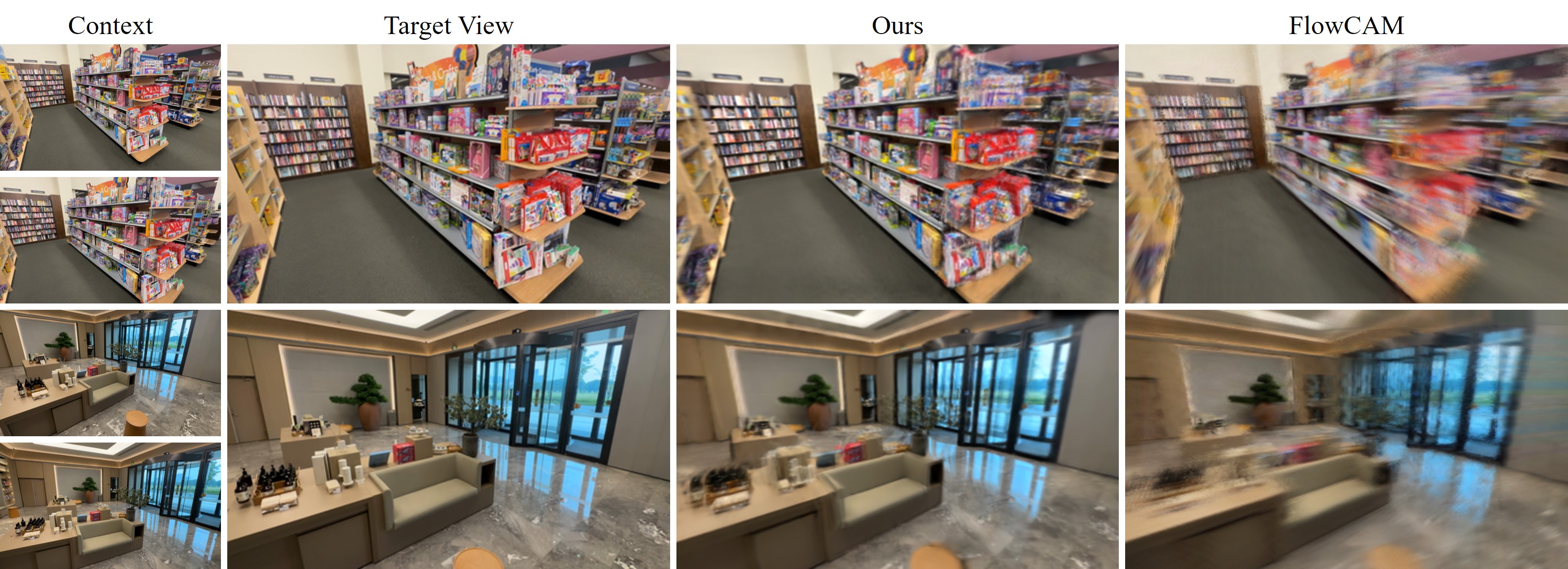}
    \vspace{-5mm}     
    \caption{Qualitative comparison of novel view synthesis on DL3DV dataset.}
    \vspace{-2mm}     
    \label{fig:qual_dl3dv}
\end{figure*}

\section{Experiments}
\label{sec:experiments}

\begin{table*}[!ht]
\renewcommand{\arraystretch}{1}
    \centering
    \resizebox{\linewidth}{!}{
\begin{tabular}{ll|cccccccccccccc}
\multicolumn{2}{c|}{}                                                       & \multicolumn{14}{c}{DL3DV}                                                                                                                                                                                                                                                                                                               \\ \hline
                                     &                                      & \multicolumn{7}{c|}{Small (10 frame)}                                                                                                                                                    & \multicolumn{7}{c}{Large (6 frame)}                                                                                                                                \\
                                     \cmidrule(lr){3-9} \cmidrule(lr){10-16}
                                     &                                      & \multicolumn{3}{c}{}                                                     & \multicolumn{2}{c}{Rotation}          & \multicolumn{2}{c|}{Translation}                           & \multicolumn{3}{c}{}                                                     & \multicolumn{2}{c}{Rotation}          & \multicolumn{2}{c}{Translation}       \\
\multicolumn{2}{l|}{Method}                                                 & PSNR $\uparrow$ & SSIM $\uparrow$ & \multicolumn{1}{l}{LPIPS $\downarrow$} & Avg.$^\circ\downarrow$ & Med.$^\circ\downarrow$ & Avg.$^\circ\downarrow$ & \multicolumn{1}{c|}{Med.$^\circ\downarrow$} & PSNR $\uparrow$ & SSIM $\uparrow$ & \multicolumn{1}{l}{LPIPS $\downarrow$} & Avg.$^\circ\downarrow$ & Med.$^\circ\downarrow$ & Avg.$^\circ\downarrow$ & Med.$^\circ\downarrow$ \\ \hline
\multicolumn{2}{l|}{FlowCAM~\cite{smith2023flowcam}} & 21.01           & 0.608           & 0.411                                & 1.138             & 1.011             & 22.385             & \multicolumn{1}{c|}{16.432}             & 23.52           & 0.710           & 0.314                                & 0.705             & 0.626             & 28.681             & 18.214             \\
\multicolumn{2}{l|}{Ours}                                                   & \best{21.91}           & \best{0.723}           & \best{0.279}                                & \best{0.985}             & \best{0.573}             & \best{9.681}             & \multicolumn{1}{c|}{\best{5.164}}             & \best{24.82}           & \best{0.822}           & \best{0.200}                                & \best{0.634}             & \best{0.256}             & \best{12.057}             & \best{6.998}            
\end{tabular}
    }
    \vspace{-2mm}          
    \caption{Quantitative results of novel view synthesis and pose estimaion on DL3DV dataset.}  
    \vspace{-3mm}          
    \label{tab:quantitative_dl3dv}
\end{table*}

\begin{table*}
\renewcommand{\arraystretch}{1}
    \centering
    \resizebox{\linewidth}{!}{
\begin{tabular}{ll|ccccccc|ccccccc}
\hline
                                     &                                      & \multicolumn{7}{c|}{RE10k $\rightarrow$ ACID}                                                                                                            & \multicolumn{7}{c}{ACID $\rightarrow$ RE10k}                                                                                                             \\
                                     \cmidrule(lr){3-9} \cmidrule(lr){10-16}
                                     &                                      & \multicolumn{3}{c}{}                                                     & \multicolumn{2}{c}{Rotation}          & \multicolumn{2}{c|}{Translation}      & \multicolumn{3}{c}{}                                                     & \multicolumn{2}{c}{Rotation}          & \multicolumn{2}{c}{Translation}       \\
\multicolumn{2}{l|}{Method}                                                 & PSNR $\uparrow$ & SSIM $\uparrow$ & \multicolumn{1}{l}{LPIPS $\downarrow$} & Avg.$^\circ\downarrow$ & Med.$^\circ\downarrow$ & Avg.$^\circ\downarrow$ & Med.$^\circ\downarrow$ & PSNR $\uparrow$ & SSIM $\uparrow$ & \multicolumn{1}{l}{LPIPS $\downarrow$} & Avg.$^\circ\downarrow$ & Med.$^\circ\downarrow$ & Avg.$^\circ\downarrow$ & Med.$^\circ\downarrow$ \\ \hline
\multicolumn{2}{l|}{FlowCAM~\cite{smith2023flowcam}} & 25.31           & 0.715           & 0.297                                & 1.607             & 1.074             & 32.316             & 21.135             & 21.13           & 0.667           & 0.329                                & 4.426             & 3.995             & 74.873             & 63.667             \\
\multicolumn{2}{l|}{CoPoNeRF~\cite{hong2023unifying}} & 23.56           & 0.683           & 0.287                                & 2.301             & 0.619             & 19.620             & 8.642             & 18.89           & 0.607           & 0.364               & 4.159             & 2.893             & 20.435             & 13.572             \\
\multicolumn{2}{l|}{Ours}                                                   & \best{26.60}           & \best{0.793}           & \best{0.206}                                & \best{1.119}             & \best{0.249}             & \best{18.607}             & \best{5.864}             & \best{21.65}           & \best{0.728}           & \best{0.242}                                & \best{1.618}             & \best{0.867}             & \best{17.993}             & \best{10.228}            
\end{tabular}
    }
    \vspace{-2mm}          
    \caption{Cross-dataset generalization. We train the models on RE10k (ACID) dataset and directly evaluate on ACID (RE10k) dataset.}
    \vspace{-4mm}
    \label{tab:quantitative_cross}
\end{table*}

\subsection{Experiment Setup}
We train and evaluate our model on three large-scale datasets: RealEstate10K (RE10k) \cite{zhou2018re10k}, ACID \cite{liu2021acid}, and DL3DV \cite{ling2024dl3dv}, which include diverse indoor and outdoor real estate videos, aerial outdoor nature scenes, and diverse real-world videos, respectively. For RE10k, we use 67,477 training and 7,289 testing scenes; for ACID, 11,075 training and 1,972 testing scenes, consistent with previous works~\cite{charatan2024pixelsplat, chen2024mvsplat}. Lastly, for DL3DV, we use subsets of the dataset amounting to 2,000 scenes (3K and 4K) for training and testing on 140 benchmark scenes, following PF3plat~\cite{hong2024pf3plat}. We assess our model's performance in reconstructing intermediate video frames between two context frames.

\noindent\textbf{Baselines.} We compare our model against existing pose-free generalizable novel view synthesis methods, including VAE~\cite{lai2021video}, DBARF~\cite{chen2023dbarf}, FlowCAM~\cite{smith2023flowcam}, and CoPoNeRF~\cite{hong2023unifying}, on two different tasks: novel view synthesis and relative camera pose estimation. We train all methods, including ours, using the same training curriculum, where the frame distance between context views increases gradually. We also provide an attribute overview in Tab.~\ref{tab:baselines}, showing the distinct features of our proposed method.

\noindent\textbf{Evaluation metrics.} For novel view synthesis, we use standard metrics: PSNR, SSIM~\cite{ssim}, and LPIPS~\cite{zhang2018lpips}. Pose estimation is evaluated based on geodesic rotation and translation angular error, following \cite{hong2023unifying}. For RE10k and ACID, we categorize test context pairs by image overlap ratios to evaluate performance across small(0.05-0.6), medium(0.6-0.8), and large(0.8+) overlap, identified by a pretrained image matching method~\cite{edstedt2024roma}. For DL3DV, overlap categories are defined by frame intervals between context images: 6 frames for large and 10 frames for small overlap.

\noindent\textbf{Implementation details.} We employ the encoder part of pretrained CroCo~\cite{weinzaepfel2023crocov2} model as our monocular encoder, which is trained in a self-supervised manner, and utilized adapter~\cite{chen2022adaptformer} block designed to efficiently adapt pretrained ViT models to downstream tasks. For the Gaussian rasterizer, we implement it using gsplat~\cite{ye2024gsplat}, an open-source library for Gaussian Splatting~\cite{kerbl20233dgs}, offering efficient computation and memory usage. We train RE10k and ACID with 256 $\times$ 256 resolution, and for DL3DV, we use 256 $\times$ 448 to accommodate the wider view in our experiments.

\vspace{-1mm}
\subsection{Results}
\noindent\textbf{Novel view synthesis.} We report quantitative results in Tab.~\ref{tab:quantitative_nvs_re10k},~\ref{tab:quantitative_nvs_acid} and qualitative results in \cref{fig:qual_re10k_acid} for RE10k and ACID datasets, while \cref{tab:quantitative_dl3dv} and \cref{fig:qual_dl3dv} for DL3DV dataset. Our method outperforms the baselines on all metrics, especially in terms of perceptual distance. These observations can be further confirmed by the rendering results that our method effectively captures fine details of 3D structure.

\begin{figure}[!ht]
    \centering
    \includegraphics[width=1.0\linewidth]{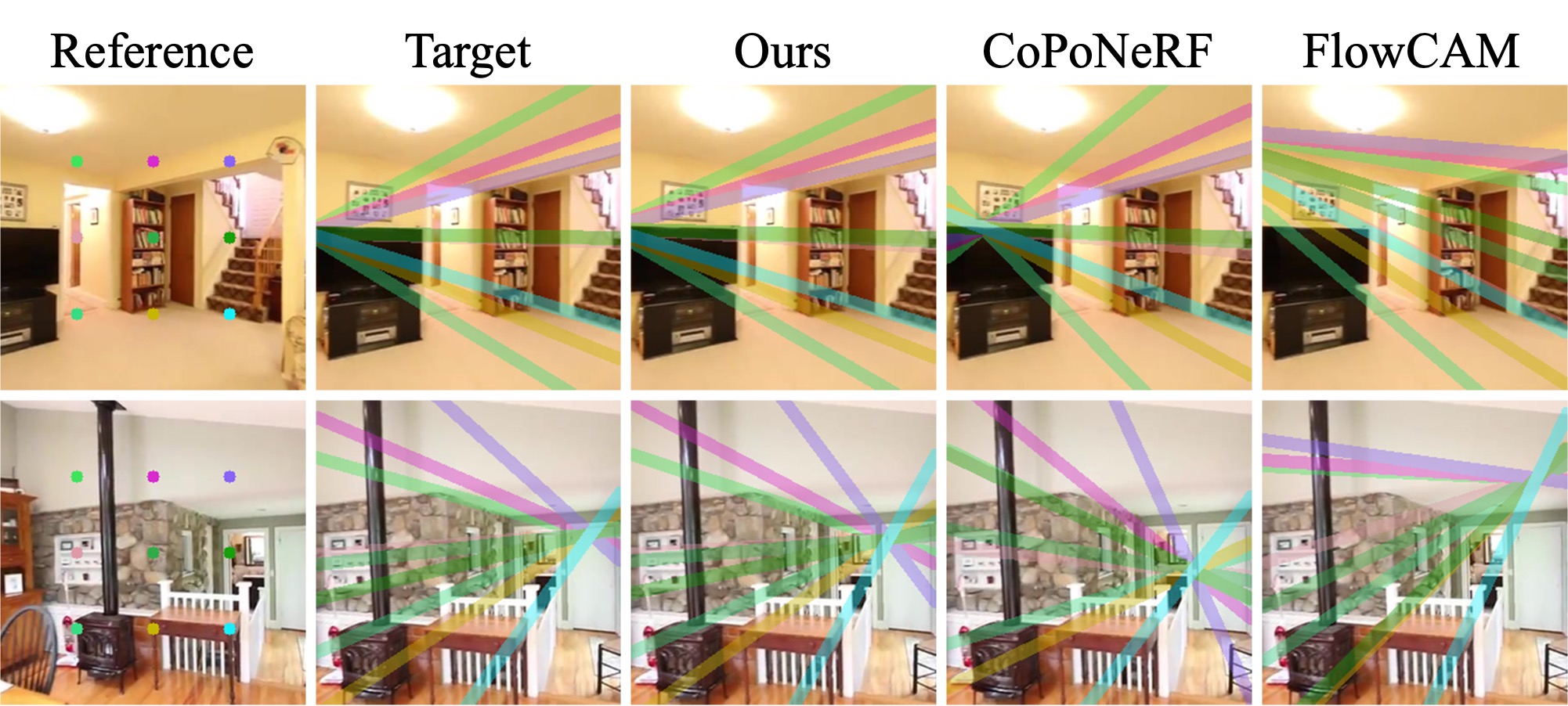}
    \vspace{-5mm}     
    \caption{Epipolar lines visualization. We draw the lines from reference to target frame using relative camera pose.}
    \vspace{-5mm}     
    \label{fig:epi}
\end{figure}

\noindent\textbf{Relative pose estimation.} Tab.~\ref{tab:quantitative_pose_re10k},~\ref{tab:quantitative_pose_acid}, and~\ref{tab:quantitative_dl3dv} present the quantitative results for camera pose estimation between two images across the datasets. Our approach consistently achieves lower errors in both average and median deviations, highlighting its accuracy and robustness. The qualitative results in \cref{fig:epi}, visualizing epipolar lines from the estimated poses also demonstrates the effectiveness of our approach in capturing accurate geometric alignments.

\noindent\textbf{Cross-Dataset Generalization.} To evaluate the generalization performance on out-of-distribution scenes, we train the models on RE10k (ACID) dataset and test them on ACID (RE10k) dataset without additional finetuning. As shown in Tab.~\ref{tab:quantitative_cross}, SelfSplat outperforms previous methods on unseen datasets, demonstrating robust generalization capabilities.

\subsection{Ablations and Analysis}
\begin{figure}[!ht]
    \centering
    \vspace{-2mm}
    \includegraphics[width=1.0\linewidth]{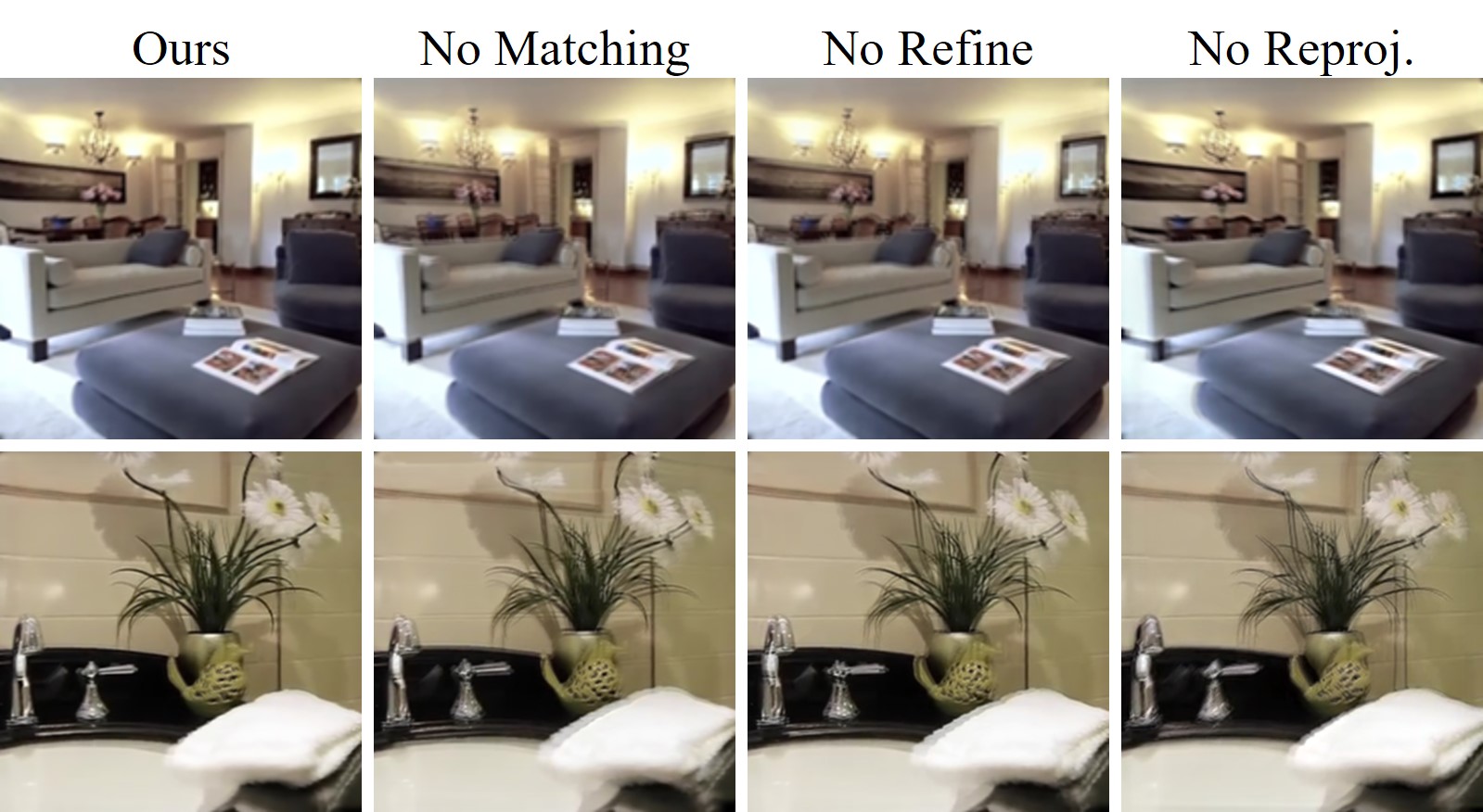}
    \vspace{-5mm}     
    \caption{Ablation studies on our proposed component.}
    \vspace{-4mm}     
    \label{fig:ablation}
\end{figure}

\vspace{-1mm}
We provide quantitative and qualitative results on ablations studies in Tab.~\ref{tab:ablations} and Fig.~\ref{fig:ablation}. All methods are trained for 50,000 iterations on RE10k dataset for a fair comparison.

\noindent\textbf{Importance of matching awareness in pose estimation.} To measure the importance of adopting cross-view features in our pose network, we conduct a study (``No Matching awareness") by removing it from the pose network. Quantitatively, it leads to a drop in pose metrics: translation error increases by 1.5 degrees, which also negatively impacts rendering scores, decreasing PSNR by 0.4 dB. These results highlight that our encoding methods with multi-view awareness help capture relationships between frames, improving both pose estimation and novel view synthesis.

\noindent\textbf{Importance of depth refinement module.} We conduct a study (``No Depth Refine") on our depth refinement module to validate its effectiveness. The results indicate a clear decline across all metrics: PSNR drops by 0.6 dB, and translation discrepancy increases by 1.1 degrees. Additionally, misalignment of overlapping Gaussians leads to degrading in visual quality, such as motion blur artifacts, demonstrating that our refinement scheme enhances the multi-view consistency of depth predictions.

\noindent\textbf{How self-supervised depth estimation method and 3D-GS representation can make reciprocal improvement?} We explore the benefits of combining self-supervised depth estimation with explicit 3D representation by comparing SelfSplat with two variants (``No Reprojection Loss", ``No Rendering Loss"). Training without reprojection loss shows a significant performance decline across all metrics, particularly a 1.5 dB drop in PSNR, indicating challenges in accurately positioning Gaussians—a crucial factor for precise 3D reconstruction and novel view synthesis. In the ``No Rendering Loss" variant, we replaced the rendered depth of the target view previously used in reprojection loss with an estimated depth map from the image using a dense prediction module. To validate the impact of incorporating 3D-GS, we also account for gradients of rotation, $R \in \mathbb{R}^{3 \times 3}$, and translation, $t \in \mathbb{R}^{3 \times 1}$, in camera poses. The rendering loss gradients with respect to translation and rotation are: 
\vspace{-3mm}
\begin{align}
&\frac{\delta \mathcal{L}_\text{ren}}{\delta t} \hspace{-1mm} = \hspace{-1mm} - \hspace{-1mm} \sum_{j}\hspace{-1mm} \frac{\delta \mathcal{L}_\text{ren}}{\delta \Tilde{\mu}_j}, \hspace{-2.5mm} &\frac{\delta \mathcal{L}_\text{ren}}{\delta R} \hspace{-1mm} = \hspace{-1mm} - \hspace{-1mm} \left[\hspace{-0.5mm} \sum_{j}\hspace{-1mm} \frac{\delta \mathcal{L}_\text{ren}}{\delta \Tilde{\mu}_j}(\mu_j \hspace{-1mm} - \hspace{-1mm} t)^{\top}\hspace{-0.5mm} \right]\hspace{-1mm} R,
\end{align}
where $\Tilde{\mu}_j$ is a splatted Gaussian in rendering viewspace. Excluding rendering loss results in degraded pose metrics, a common issue in self-supervised depth estimation methods with limited image overlap. Our framework effectively addresses this by combining explicit 3D-GS representation with rendering loss, improving depth and pose estimation.

\begin{table}[!ht]
    \centering
    \resizebox{\linewidth}{!}{
\begin{tabular}{ll|ccccc}
\toprule
\multicolumn{2}{l|}{Method}                & PSNR $\uparrow$ & SSIM $\uparrow$ & LPIPS $\downarrow$ & Rot. Avg.$^\circ\downarrow$ & Trans. Avg.$^\circ\downarrow$ \\ \midrule
\multicolumn{2}{l|}{Ours}                  & \best{22.65}           & \best{0.764}           & \best{0.222}              & \best{1.036}                 & \best{13.705}                    \\
\multicolumn{2}{l|}{No Matching Network} & 22.21           & 0.735           & 0.241              & 1.308                 & 15.171                    \\
\multicolumn{2}{l|}{No Depth Refine}       & 22.05           & 0.744           & 0.224              & 1.064                 & 14.8                    \\
\multicolumn{2}{l|}{No Reprojection Loss}  & 21.12           & 0.672           & 0.293              & 2.236                 & 28.584                    \\
\multicolumn{2}{l|}{No Rendering Loss}     & -               & -               & -                  & 8.581                 & 64.436                    \\ \bottomrule
\end{tabular}
    }
    \vspace{-1mm}    
    \caption{Ablations. Our methods achieves better alignment of 3D Gaussians, with accurate pose and consistent depth estimations.}    
    \vspace{-3mm}    
    \label{tab:ablations}
\end{table}
\vspace{-2mm}
\section{Conclusion}
\label{sec:conclusion}
\vspace{-1mm}
We present SelfSplat, a pose-free generalizable 3D Gaussian Splatting model that does not require pretrained 3D priors or an additional fine-tuning stage. Our method effectively integrates a 3D-GS representation with self-supervised depth estimation techniques to recover 3D geometry and appearance from unposed monocular videos. We conduct extensive experiments on diverse real-world datasets to demonstrate its effectiveness, showcasing its ability to produce photorealistic novel view synthesis and accurate camera pose estimation. We believe that SelfSplat represents a significant step forward in 3D representation learning, offering a robust solution for various applications.

\appendix

\twocolumn[
    \centering
    \Large
    \textbf{\thetitle}\\
    \vspace{0.5em}Supplementary Material \\
    \vspace{1.0em}
] %

\section{Additional Details}
\subsection{Architectural Details}
\begin{figure*}[!ht]
    \centering
    \includegraphics[width=0.9\textwidth]{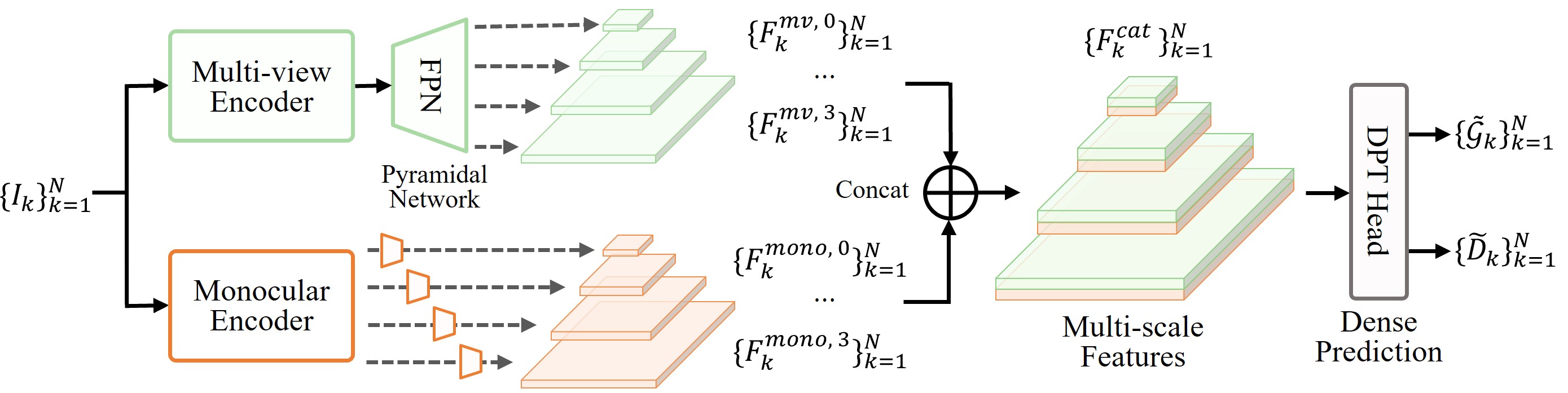}
    \caption{Detailed 3D Gaussian prediction architecture. This module takes only context images as input.}
    \vspace*{-2mm}
    \label{fig:dpt}
\end{figure*}

For the prediction of 3D Gaussians~\cite{kerbl20233dgs}, we utilize the monocular, multi-view encoder and the fusion block. Unlike previous methods that utilize DepthAnything~\cite{yang2024depthanything} as a monocular encoder~\cite{zhang2024transplat, xu2024depthsplat} or UniMatch~\cite{xu2023unifying} as a multi-view encoder~\cite{chen2024mvsplat, sc_depthv3}, we only employ the encoder part of Croco~\cite{weinzaepfel2022crocov1} as our monocular encoder which is trained in a fully self-supervised manner. For the multi-view encoder, we adopt the backbone of~\cite{xu2023unifying} with randomly initialized weights. Then, we unify features from monocular and multi-view encoders using DPT~\cite{ranftl2021dpt} block. For a detailed architecture for the fusion module, see Fig.~\ref{fig:dpt}.

\subsection{Implementation Details}
For our monocular encoder, we utilized Adapter~\cite{chen2022adaptformer}, which keeps the model parameters frozen while training additional residual networks for each layer. Specifically, a residual MLP block, comprising a down-projection layer and an up-projection layer, is introduced within each layer of the transformer encoder. Considering the channel dimension of the original encoder, $C^\text{mono} = 1024$, we set the low rank hidden dimension of AdaptMLP, $C^\text{adapt} = 32$, to efficiently reduce computational overhead while maintaining sufficient capacity for adaptation.

For 3D Gaussian primitives, we set the order of spherical harmonics expansion to 1, enabling the representation to extend beyond the Lambertian color model. When warping the color model from each frame’s local coordinate system into an integrated global space which requires the Wigner matrices in general case, we simplify the rotation of the first level of spherical harmonics, $Y_1(r_d) \hspace{-1mm} = \hspace{-1mm} [Y_1^{-1}(r_d), Y_1^0(r_d), Y_1^1(r_d)]$, as follows:
$$
Y_1(r_d) = \sqrt{\frac{3}{4 \pi}} \Pi \hspace{0.5mm} r_d,
\quad
\Pi =
\begin{bmatrix}
  0 & 1 & 0 \\
  0 & 0 & 1 \\
  1 & 0 & 0 \\
\end{bmatrix},
$$
where $r_d\in\mathbb{S}^2$ is the viewing direction derived from the estimated camera poses. We adopt this warping protocol from Splatter Image~\cite{szymanowicz2024splatter} which is a pose-required generalizable 3D reconstruction model using 3D Gaussian Splatting.

\subsection{Training Details}
We train all baseline models, including ours, using custom data loaders. For RealEstate10K~\cite{zhou2018re10k} (RE10k) and ACID~\cite{liu2021acid} datasets, the distance between context frames is progressively increased from 5 to 25, and target frames are randomly selected between the context frames within this range. Each model is trained for 200K iterations and for baselines we used the default hyperparameter settings provided by the respective authors. The only exception is DBARF~\cite{chen2023dbarf}, which is trained for 400K iterations due to its official implementation supporting only a batch size of one. We provide our detailed training hyperparameters in Tab.~\ref{tab:training_config} and we train our model on a singe H100 GPU, which takes approximately for 3 days. For the experiment on DL3DV~\cite{ling2024dl3dv} dataset, we initialize the model with pretrained weights from RE10k dataset and train it for 50K iterations on a single H100 GPU with a batch size of 6.  The distance between context frames is gradually increased from 2 to 10. This procedure is applied to FlowCAM~\cite{smith2023flowcam} in the same way which is the baseline model on DL3DV dataset.

For VAE~\cite{lai2021video}, which was initially designed for novel view synthesis from a single image, we modify its architecture following the approach in~\cite{yu2021pixelnerf} to handle multi-view input images. Specifically, we employ two separate encoders and use their mean output as the input to the decoder which synthesize novel view images. All other hyperparameters remain the same as the official implementation.

\begin{table}[!ht]
    \centering
    \resizebox{0.6\columnwidth}{!}{
\begin{tabular}{lc}
\hline
\multicolumn{2}{c}{SelfSplat}                    \\ \hline
Config        & Value                               \\ \hline
optimizer     & Adam~\cite{kingma2014adam} \\
scheduler     & Linear                              \\
learning rate & 1e-4                                \\
gradient clipping & 0.5                             \\
batch size & 12                                \\
total iters.  & 200,000                               \\ 
warmup iters.  & 2,000                               \\ \hline
\end{tabular}
    }
    \caption{Training hyperparameters.}    
    \label{tab:training_config}    
\end{table}

\vspace{-3mm}

\subsection{Evaluation Details}
During the evaluation on RE10k and ACID datasets, we set the interval between context frames to 40 and select the middle frame as the target view point. This target frame is used as the ground truth for metric evaluations in novel view synthesis and camera pose estimation. For the overlap categories, we utilize the pretrained feature matching model, RoMa~\cite{edstedt2024roma}, to estimate the overlap ratios between the first context frame and the target frame. 

For RE10k dataset, the split proportions are 18.26\% for large, 60.56\% for medium, and 21.17\% for small categories. In ACID dataset~\cite{liu2021acid}, the proportions are 33.05\% for large, 41.15\% for medium, and 25.80\% for small.

\section{Additional Experiment Analysis}
\subsection{Inference Cost}

We report the memory and time consumption required to synthesize a single 256 × 256 image during the inference stage in Table~\ref{tab:resource}. Memory usage is measured as the peak memory during inference, while the number of rays per batch is adjusted if necessary. Except for VAE~\cite{lai2021video}, which generates novel view images without rendering operations (utilize 2D CNN blocks) and thus fail to reconstruct interpretable 3D scene representations, our method achieves significantly lower memory usage and faster rendering speed with explicit 3D representations, demonstrating its efficiency and practical usage in real-world scenarios.

\begin{table}[!ht]
\centering
\resizebox{0.7\columnwidth}{!}{
\begin{tabular}{l|cc}
\toprule 
Method   & \multicolumn{1}{l}{Mem. (GB)} & \multicolumn{1}{l}{Time (s)} \\ \hline
VAE~\cite{lai2021video}      & 4.694                         & \best{0.0003}                       \\
DBARF~\cite{chen2023dbarf}    & 2.079                         & 0.254                        \\
FlowCAM~\cite{smith2023flowcam}  & 16.644                        & 0.801                        \\
CoPoNeRF~\cite{hong2023unifying} & 16.802                        & 5.624                        \\
Ours     & \best{1.795}                         & 0.002        \\ \bottomrule              
\end{tabular}}
\caption{Memory and time consumption analysis. All baselines including ours are measured on a single NVIDIA RTX 4090 GPU.}
\label{tab:resource}   
\end{table}

\subsection{Using N Context Views}
\begin{figure}[!ht]
    \centering
    \includegraphics[width=0.75\linewidth]{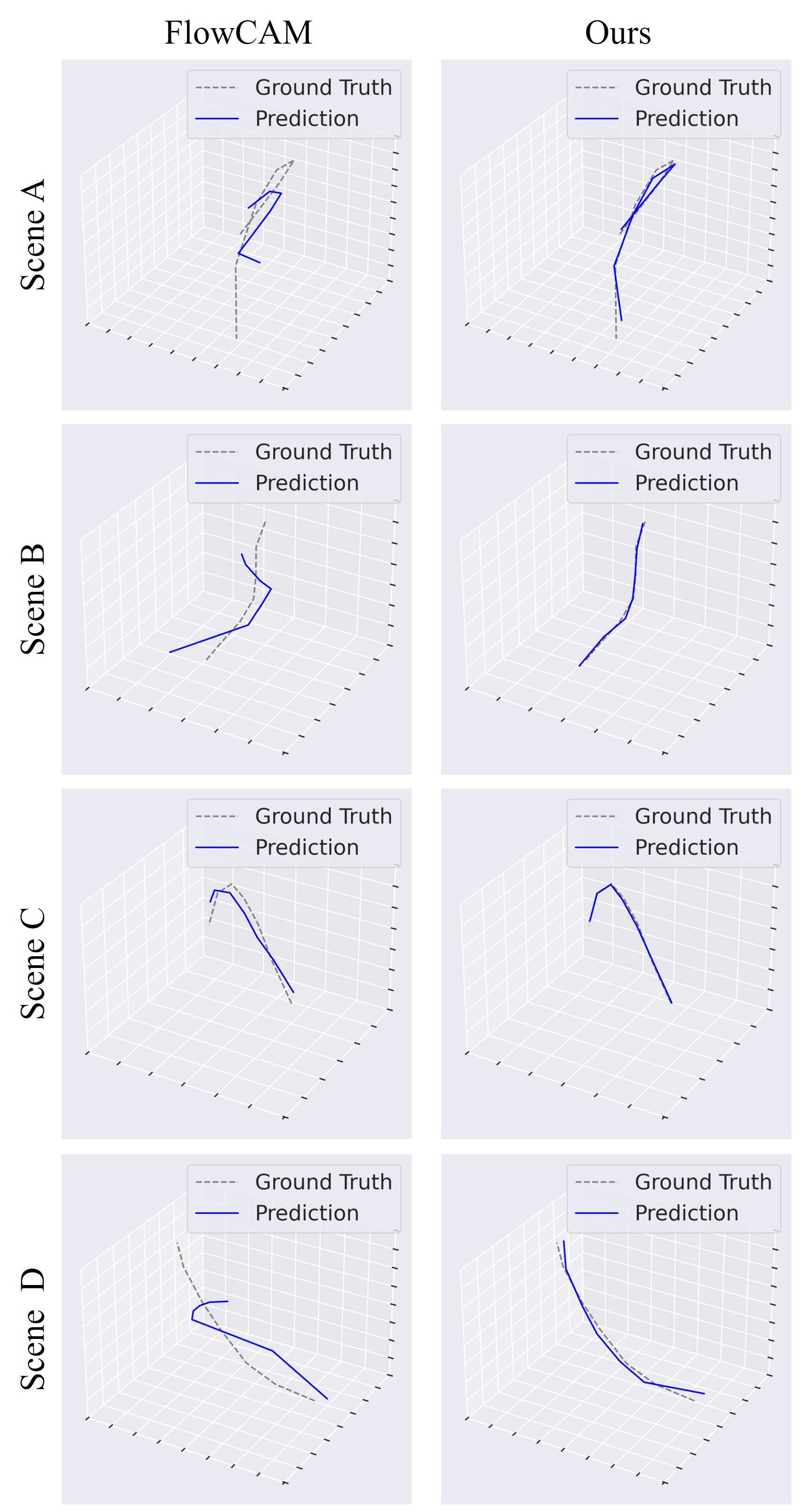}
    
    \caption{Visualization of camera trajectory on RE10k dataset. Construction of trajectory only consider the translation part of the estimated camera poses.}
    \vspace{-3mm}
    \label{fig:cam_traj}
\end{figure}

We further evaluate the model's performance across various numbers of input views, considering its practical application where more than two views are commonly used. The total number of frames is evenly divided based on the number of context views, and target frames are sampled between the context frames. Additionally, we generate a camera trajectory using the selected view points (context and target), and the Absolute Trajectory Error (ATE) is measured to validate the accuracy of the reconstructed camera path. We evaluate on RE10k dataset with 3 context views (80 frames) and 4 context views (120 frames) settings. As shown in Tab.~\ref{tab:multiview} and Fig.~\ref{fig:cam_traj}, our method demonstrates superior performance in both novel view synthesis and camera trajectory estimation, as well as its ability to scale effectively with multiple input views and estimations over extended ranges without any further finetuning.

\begin{table}[!ht]
\centering
\resizebox{1.0\columnwidth}{!}{
\begin{tabular}{l|cccc|cccc}
        & \multicolumn{4}{c|}{3 views (80 frames)} & \multicolumn{4}{c}{4 views (120 frames)} \\ \cmidrule(lr){2-5} \cmidrule(lr){6-9} 
Method  & PSNR$\uparrow$     & SSIM$\uparrow$      & LPIPS$\downarrow$    & ATE$\downarrow$    & PSNR$\uparrow$     & SSIM$\uparrow$     & LPIPS$\downarrow$     & ATE$\downarrow$    \\ \hline
FlowCAM~\cite{smith2023flowcam} & 19.75    & 0.630    & 0.412         & 0.048       & 18.91         & 0.606         & 0.449          & 0.081       \\
Ours    & \best{21.12}    & \best{0.761}     & \best{0.241}         & \best{0.031}       & \best{19.52}         & \best{0.717}         & \best{0.283}          & \best{0.053}      
\end{tabular}}

\caption{Quantitative results of using different numbers of context views on RE10k dataset.}
\vspace{-2mm}
\label{tab:multiview}  
\end{table}
\vspace{-3mm}

\subsection{Additional Comparison}
\begin{figure}[!ht]
    \centering
    \includegraphics[width=1.0\linewidth]{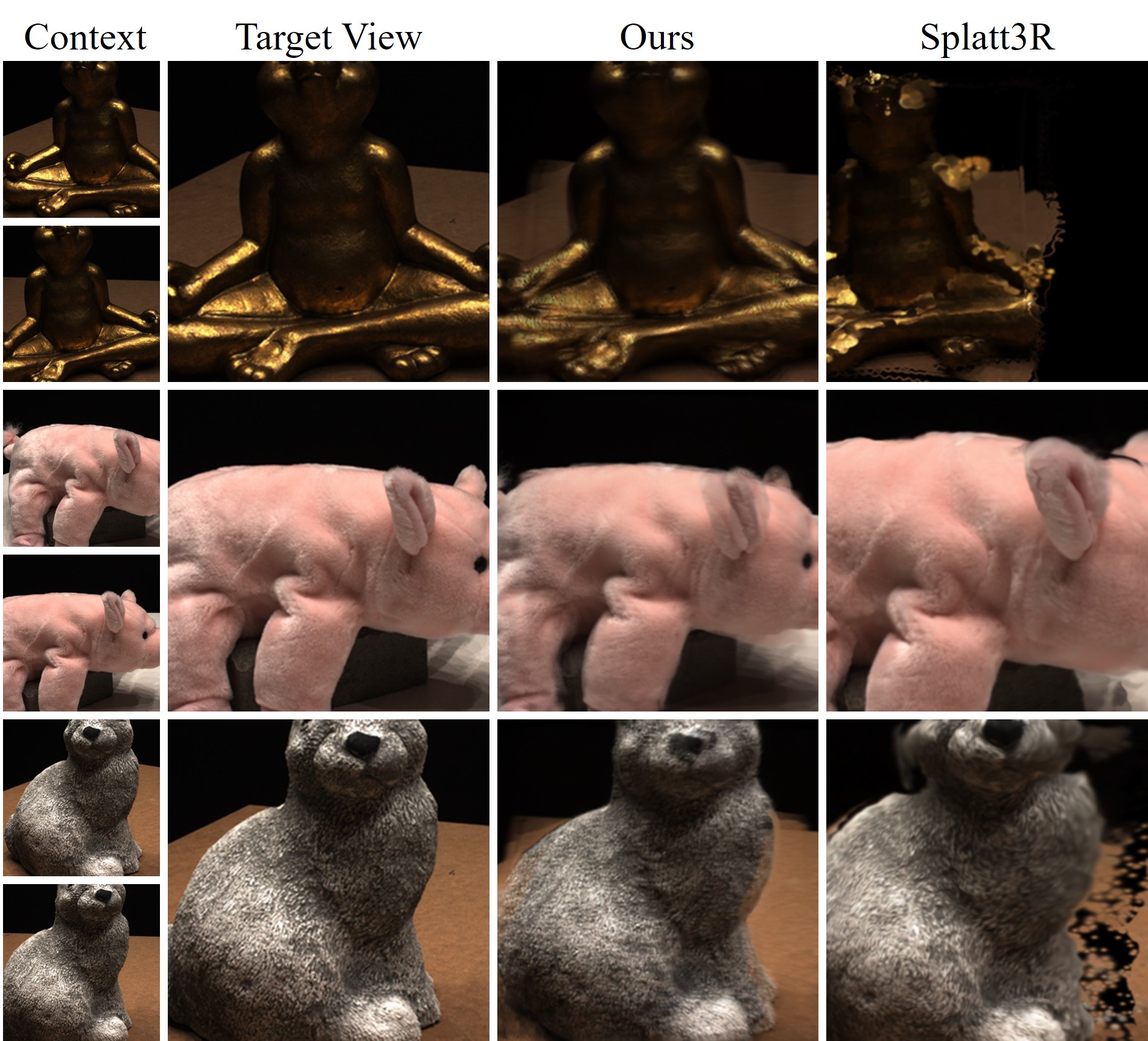}
    \caption{Qualitative comparison of novel view synthesis on DTU dataset.}
    \vspace{-3mm}
    \label{fig:dtu}
\end{figure}

For the reader's reference, we provide a comparison with Splatt3R~\cite{smart2024splatt3r}, which is also a pose-free, feed-forward Gaussian Splatting method for 3D reconstruction and novel
view synthesis from stereo pairs. We omitted this model in the main paper because it requires ground-truth depth and camera pose annotations during training, which are not available in the datasets we used: RE10k, ACID~\cite{liu2021acid}, and DL3DV~\cite{ling2024dl3dv}. Acknowledging the differences in training data—Splatt3R was trained on ScanNet++~\cite{dai2017scannet}, whereas our model was trained on RE10k—we evaluate them on the DTU~\cite{jensen2014dtu} dataset, which is an out-of-distribution dataset for both. As shown in Tab.~\ref{tab:dtu} and Fig.~\ref{fig:dtu}, our method achieves better performance than the baseline in both evaluation metrics and visual quality, and also outperforms pixelSplat~\cite{charatan2024pixelsplat} which is a pose-required method in training and evaluation stage. The main reason Splatt3R cannot estimate a consistent scene scale is its reliance on a fixed pretrained MASt3R~\cite{leroy2024grounding} model, which is trained using metric camera poses, and difference between estimated intrinsic parameters and ground truth intrinsic parameters. Thus, using the DTU dataset, which consists of unseen novel scenes, they fail to align consistent 3D Gaussians.

\begin{table}[!ht]
\centering
\resizebox{0.9\columnwidth}{!}{
\begin{tabular}{ll|ccc}
Training Data & Method     & PSNR$\uparrow$                 & SSIM$\uparrow$                 & LPIPS$\downarrow$                \\ \hline
ScanNet++~\cite{dai2017scannet}     & Splatt3R~\cite{smart2024splatt3r}   & 10.24                & 0.295               & 0.629                     \\ \hline
              & \color{gray}pixelSplat~\cite{charatan2024pixelsplat} & \multicolumn{1}{c}{\color{gray}12.89} & \multicolumn{1}{c}{\color{gray}0.382} & \multicolumn{1}{c}{\color{gray}0.560} \\
RE10k~\cite{zhou2018re10k} & \color{gray}MVSplat~\cite{chen2024mvsplat}    & \multicolumn{1}{c}{\color{gray}13.94} &  \multicolumn{1}{c}{\color{gray}0.473} & \multicolumn{1}{c}{\color{gray}0.385} \\
              & Ours       & \best{13.14}                & \best{0.425}                &  \best{0.448}                   
\end{tabular}}
\caption{Quantitative results of novel view synthesis on DTU dataset. While pixelSplat~\cite{charatan2024pixelsplat} and MVSplat~\cite{chen2024mvsplat} are pose-required methods, we include them for the reader's reference.}
\label{tab:dtu}  
\end{table}
\vspace{-5mm}

\subsection{Baseline Comparisons}
We provide additional baseline results on cross-dataset generalization in Tab.~\ref{tab:sup_cross}.
\begin{table}[!ht]
\renewcommand{\arraystretch}{0.95}
    \centering
    \resizebox{1.0\linewidth}{!}{
\begin{tabular}{ll|cccc|cccc}
\hline
                                     &                                      & \multicolumn{4}{c|}{RE10k $\rightarrow$ ACID}                                                                                                            & \multicolumn{4}{c}{ACID $\rightarrow$ RE10k}                                                                                                             \\
\multicolumn{2}{l|}{Method}                                                 & PSNR $\uparrow$ & \multicolumn{1}{l}{SSIM $\uparrow$} & Rot.$^\circ\downarrow$ & Trans.$^\circ\downarrow$ &  PSNR $\uparrow$ & \multicolumn{1}{l}{SSIM $\uparrow$} & Rot.$^\circ\downarrow$ &  Trans.$^\circ\downarrow$  \\ \hline
\multicolumn{2}{l|}{VAE} & 23.67                 & 0.649                                & 1.619                 & 117.721                 & 18.98                 & 0.537                                & 3.744                   & 65.884             \\
\multicolumn{2}{l|}{DBARF} & 23.56                 & 0.644                                & 1.772            & 51.969                & 12.60           & 0.502               & 3.306          & 47.851          \\
\multicolumn{2}{l|}{Ours}                                                   & \best{26.60}               & \best{0.793}                                & \best{1.119}          & \best{18.607}                & \best{21.65}                 & \best{0.728}                                & \best{1.618}                    & \best{17.993}      
\end{tabular}}
\vspace{-2mm}
\caption{Additional comparison on cross-dataset generalization.}
\vspace{-4mm}
\label{tab:sup_cross}
\end{table}

\subsection{Additional Ablation and Analysis}
We provide additional ablation studies and analyses, focusing on our encoder module. All methods are trained on RE10k~\cite{zhou2018re10k} for 50k iterations, following the same procedure as in the main paper. As shown in Tab.~\ref{tab:sup_ab}, our feature fusion module with CroCo~\cite{weinzaepfel2022crocov1} initialization shows superior results in evaluation metrics. 
\begin{table}[!ht]
\centering
\resizebox{1.0\linewidth}{!}{
\begin{tabular}{ll|ccccc}
\toprule
\multicolumn{2}{l|}{Method}                & PSNR $\uparrow$ & SSIM $\uparrow$ & LPIPS $\downarrow$ & Rot. Avg.$^\circ\downarrow$ & Trans. Avg.$^\circ\downarrow$ \\ \midrule
\multicolumn{2}{l|}{Ours}                  & \best{22.65}           & \best{0.764}           & \best{0.222}              & \best{1.036}                 & \best{13.705}                \\
\multicolumn{2}{l|}{No CroCo~\cite{weinzaepfel2022crocov1} Init} & 22.15           & 0.738           & 0.240              & 1.091                 & 14.042                    \\
\multicolumn{2}{l|}{No Monocular Encoder}       & 21.88           & 0.733           & 0.247              & 1.394                 & 17.125                    \\
\multicolumn{2}{l|}{No Multi-view Encoder}  & 21.47           & 0.731           & 0.243              & 1.233                 & 16.327                                 
\\ \bottomrule
\end{tabular}} 
\vspace{-2mm}
\caption{Ablation studies on the encoder module design.}
\vspace{-3mm}
\label{tab:sup_ab}  
\end{table}

\noindent\textbf{Pretrained weight.} Since our goal is to use only unposed raw video datasets without 3D priors, we utilized CroCo, trained in a self-supervised manner. While DUSt3R~\cite{wang2024dust3r} or MASt3R~\cite{leroy2024grounding} pre-trained weight could enhance performance, we focus on demonstrating that 3D foundation models can be trained without costly 3D annotations.

\subsection{Architectural and Evaluation Design} 
We designed our evaluation protocol assuming that there are no given poses, so we made a separate pose block (context and target) and a Gaussian branch (only context) independently. Thus, target images are used to estimate camera poses for following novel view synthesis evaluations. All baselines follow this protocol in their original implementations, except for CoPoNeRF~\cite{hong2023unifying} which utilizes given camera poses, so we substitute these poses in CoPoNeRF with estimated ones for a fair comparison. 

\subsection{Depth Visualization}

\begin{figure}[!ht]
    \centering
    \includegraphics[width=1.0\linewidth]{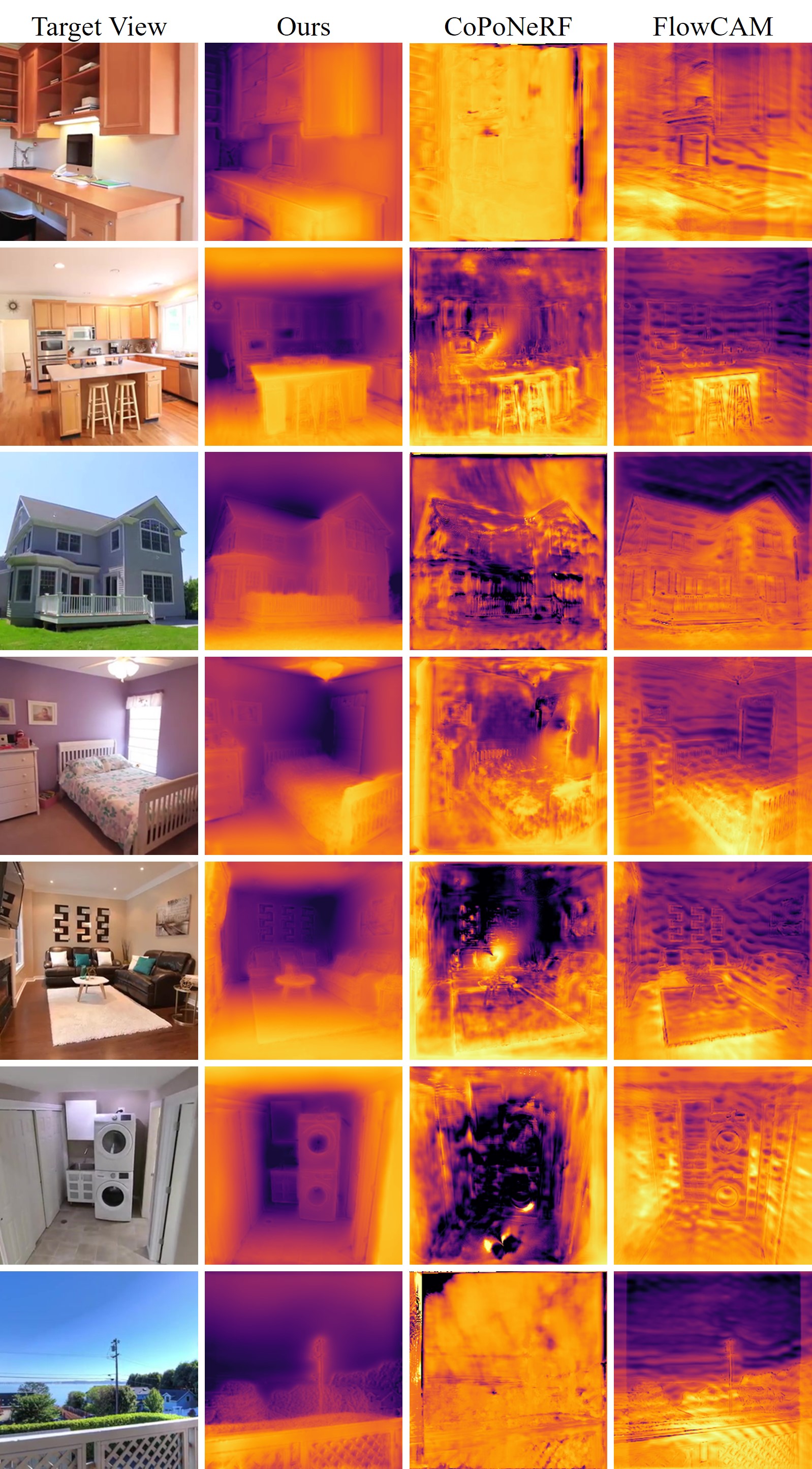}
    \vspace{-2mm}
    \caption{Qualitative comparison of depth visualization on RE10k dataset. Depth maps are obtained following the rendering process.}
    \vspace{-4mm}
    \label{fig:depth}
\end{figure}

We also provide the visualization of depth maps generated through rendering, which is essential for producing interpretable 3D representations. By comparing the results of our method with previous approaches, as shown in Fig.~\ref{fig:depth}, SelfSplat demonstrates robust and reliable depth maps derived from 3D scene structures.

\section{Limitations} While we demonstrate high-quality 3D geometry estimation in this work, the current framework still possesses limitations. First, further technical improvements are needed to support wider baseline scenarios, such as a 360$^\circ$ scene reconstruction from unposed images in a single forward pass. Second, our framework struggles with dynamic scenes where both camera and object motion are present. Addressing these complex scenarios may benefit from incorporating multi-modal priors~\cite{rombach2022high, sun2023emu} for robust and consistent alignment across wide and dynamic scene space.

\section{Additional Results}

We provide additional results on the following pages including novel view synthesis and epipolar line visualizations.

\clearpage

\begin{figure*}[!ht]
    \centering
    \includegraphics[width=1.0\textwidth]{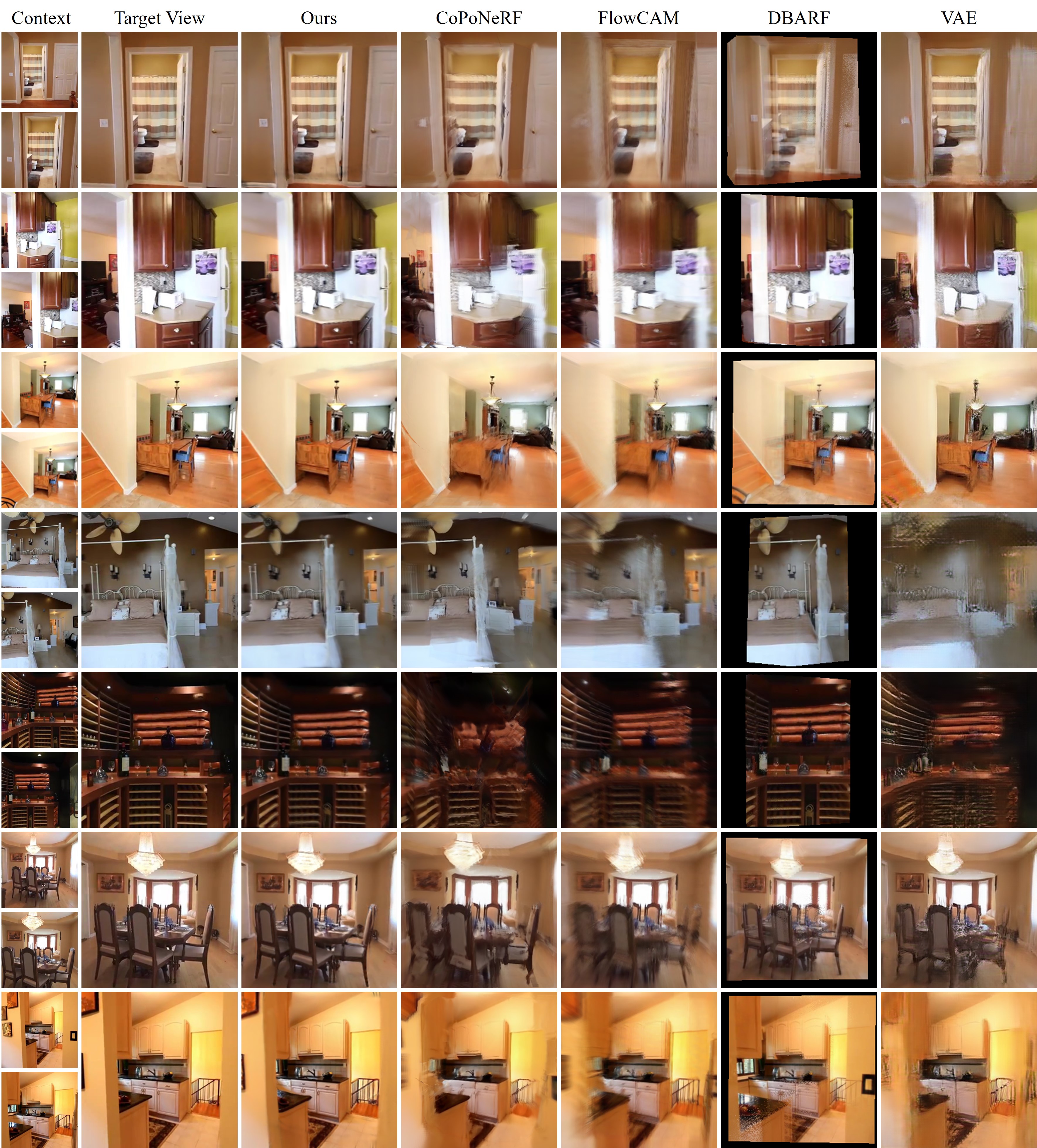}
    \caption{Qualitative comparison of novel view synthesis on RE10k dataset.}
    \label{fig:qual_re10k}
\end{figure*}

\begin{figure*}[!ht]
    \centering
    \includegraphics[width=1.0\textwidth]{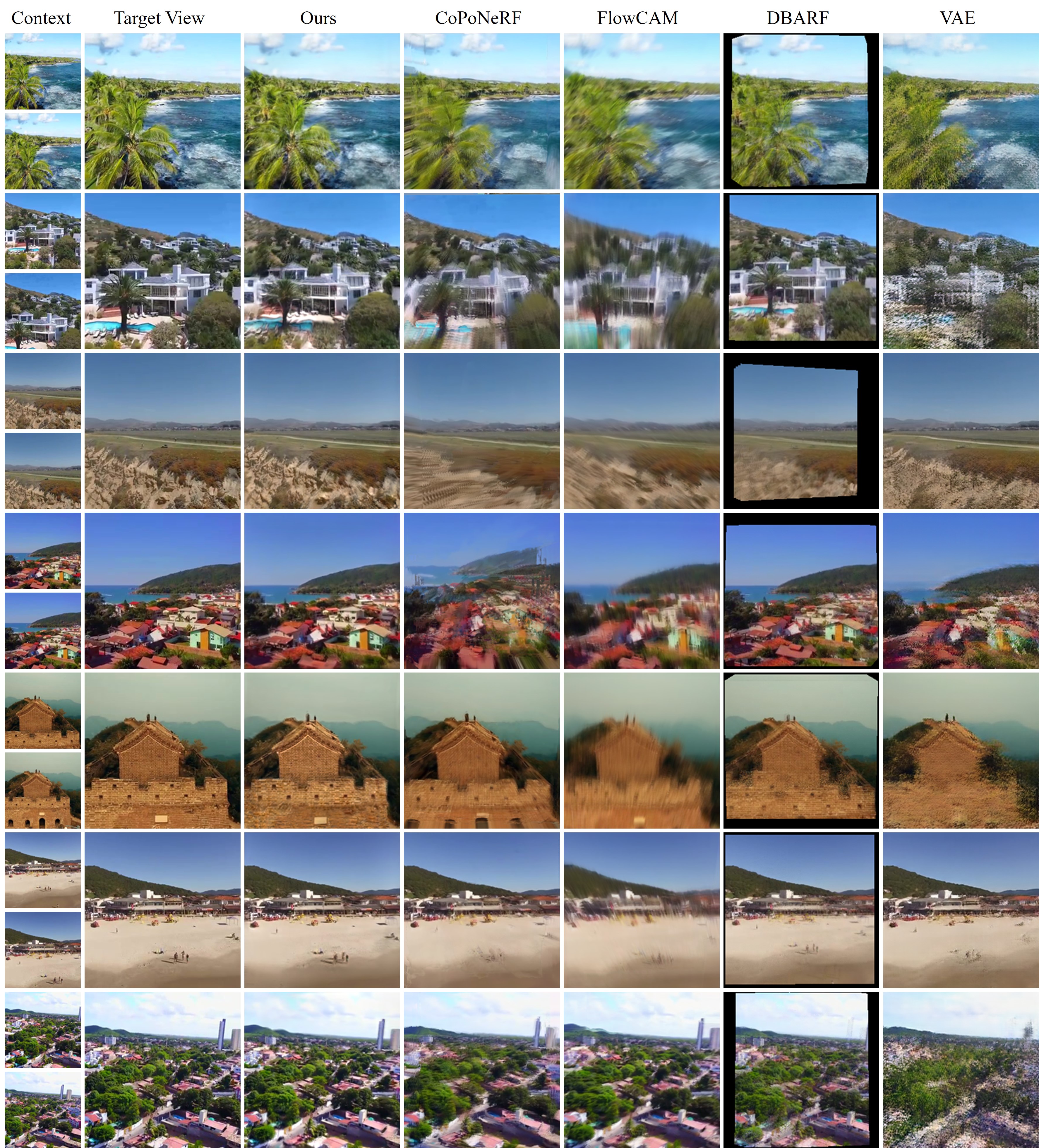}
    \caption{Qualitative comparison of novel view synthesis on ACID dataset.}
    \label{fig:qual_acid}
\end{figure*}

\begin{figure*}[!ht]
    \centering
    \includegraphics[width=1.0\textwidth]{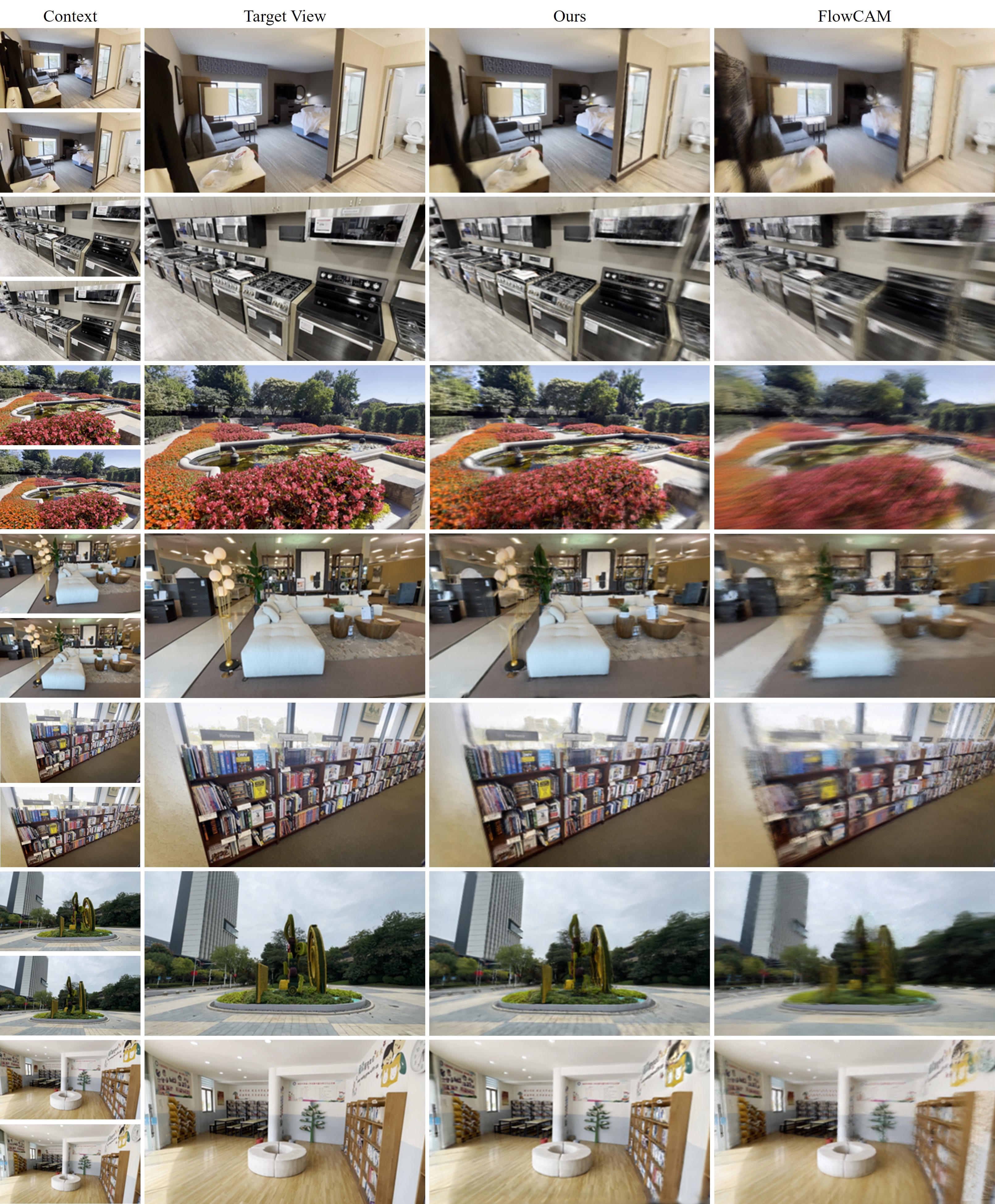}
    \caption{Qualitative comparison of novel view synthesis on DL3DV dataset.}
    \label{fig:sub_qual_dl3dv}
\end{figure*}

\begin{figure*}[!ht]
    \centering
    \includegraphics[width=1.0\textwidth]{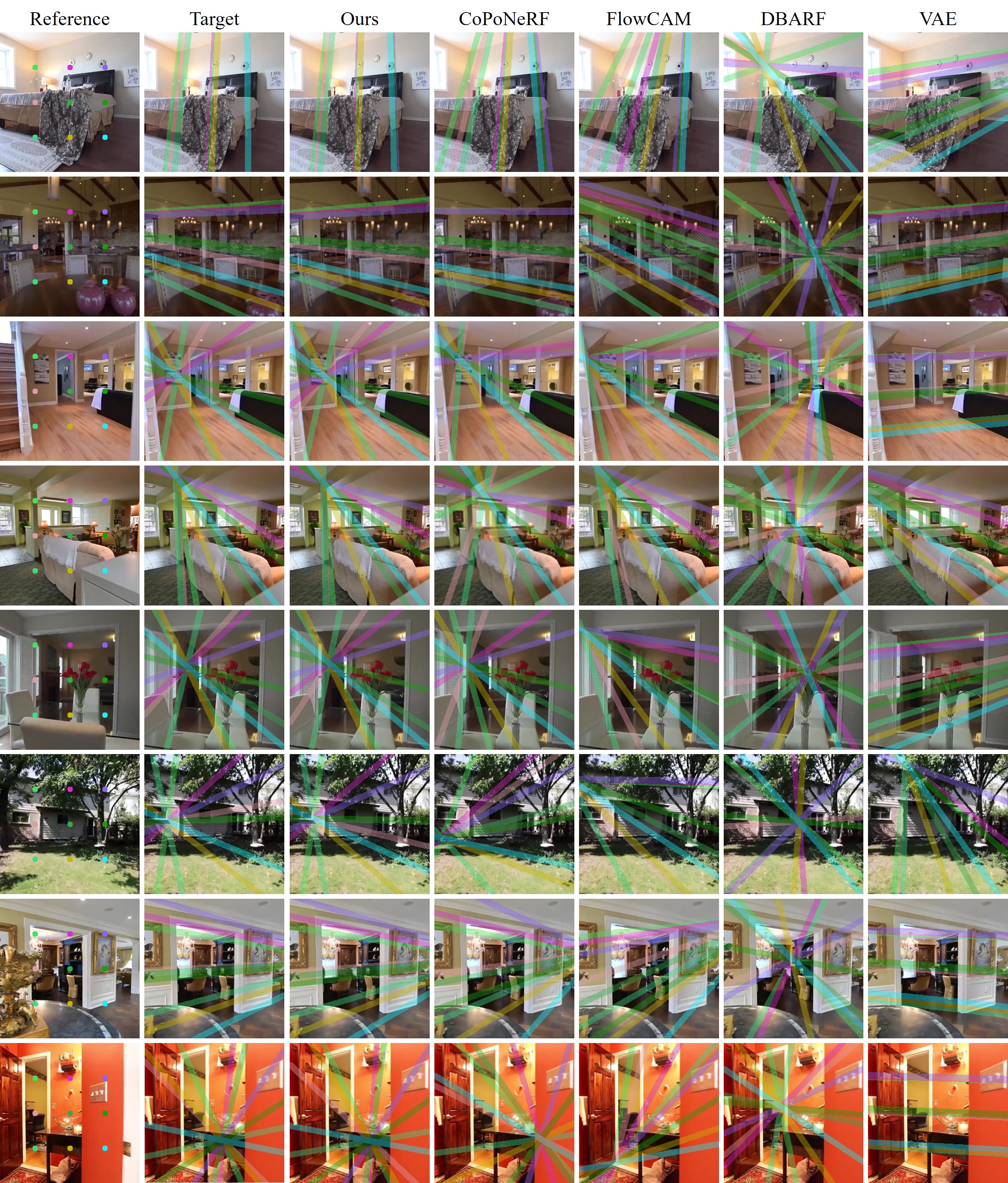}
    \caption{Epipolar lines visualization on RE10k dataset. We draw the lines from reference to target frame using relative camera pose.}
    \label{fig:qual_epi}
\end{figure*}

\clearpage

{
    \small
    \bibliographystyle{ieeenat_fullname}
    \bibliography{main}
}


\end{document}